\documentclass[11pt]{article}

\usepackage[T1]{fontenc}
\usepackage{lmodern}
\usepackage[margin=1in]{geometry}
\usepackage{amsmath,amssymb}
\usepackage{mathtools}
\usepackage{array,booktabs,tabularx}
\usepackage{adjustbox}
\usepackage{graphicx}
\usepackage{float}
\usepackage{xcolor}
\usepackage{microtype}
\usepackage[numbers,sort&compress]{natbib}
\usepackage{hyperref}
\usepackage{listings}
\usepackage{tikz}
\usepackage{placeins}
\usepackage{needspace}
\usepackage{graphcore_report}
\usetikzlibrary{arrows.meta,positioning,fit}

\newcolumntype{Y}{>{\raggedright\arraybackslash}X}
\newcommand{\projectrepo}{\href{https://github.com/MrHuff/fp4-fa4}{\texttt{github.com/MrHuff/fp4-fa4}}}

\floatstyle{ruled}
\newfloat{algorithm}{tbp}{loa}
\floatname{algorithm}{Algorithm}
\restylefloat{algorithm}

\title{Hardware-Aware FP4 FlashAttention-4}
\reportsubtitle{Full-FP4 forward inference, quantized causal backward, and transformer training}
\author{Robert Hu}
\date{September 2026}
\reportcontact{%
  Correspondence: \href{mailto:robert.stats.hu@gmail.com}{\texttt{robert.stats.hu@gmail.com}}}
\hypersetup{
  colorlinks=true,
  linkcolor=GraphcoreInk,
  citecolor=GraphcoreCoralDark,
  urlcolor=GraphcoreCoralDark,
  pdftitle={Hardware-Aware FP4 FlashAttention-4},
  pdfauthor={Robert Hu},
  pdfsubject={Hardware-aware FP4 FlashAttention-4 on NVIDIA GB200 and B300},
  pdfkeywords={FlashAttention-4, FP4, NVFP4, MXFP4, Blackwell, tensor memory, softmax}
}

\definecolor{scoreblue}{HTML}{2A6FBB}
\definecolor{outputgreen}{HTML}{28865D}
\definecolor{probred}{HTML}{C83E4D}
\definecolor{scaleyellow}{HTML}{D99B22}
\definecolor{inkgray}{HTML}{3C4650}
\definecolor{lightgray}{HTML}{EDF0F2}

\newcommand{\E}{\mathbb{E}}

\newcommand{\Qe}{Q_{\mathrm{E2M1}}}
\newcommand{\softmax}{\operatorname{softmax}}
\newcommand{\code}[1]{\texttt{#1}}
\newcommand{\us}{\ensuremath{\,\mu\mathrm{s}}}

\newcommand{\tcgen}{\texttt{tcgen05}}
\newcommand{\reportplot}[2][0.20\textheight]{%
  \IfFileExists{#2}{%
    \includegraphics[width=\textwidth]{#2}%
  }{%
    \fbox{\parbox[c][#1][c]{0.94\textwidth}{\centering
      \color{inkgray}\bfseries Plot placeholder\par\medskip
      \normalfont\small Expected asset:\par
      \texttt{\detokenize{#2}}}}%
  }%
}

\newcommand{\DownstreamClassificationSamples}{2272}
\newcommand{\DownstreamFastPredictionChanges}{32}
\newcommand{\DownstreamFastLowMarginChanges}{31}

\newcommand{\HaoGridFastPeakTflops}{2998}
\newcommand{\HaoGridFastGeoSpeedup}{2.023}
\newcommand{\HaoGridFastMeanCosine}{0.943789}
\newcommand{\HaoGridFastMeanRelLTwo}{0.336602}
\newcommand{\HaoGridAccuratePeakTflops}{2416}
\newcommand{\HaoGridAccurateGeoSpeedup}{1.669}
\newcommand{\HaoGridAccurateMeanCosine}{0.951669}
\newcommand{\HaoGridAccurateMeanRelLTwo}{0.327225}

\newcommand{\HaoPublishedMeanCosine}{0.9899}

\newcommand{\BThreeDsixtyfourMinSpeedup}{1.124}
\newcommand{\BThreeDsixtyfourMaxSpeedup}{1.284}
\newcommand{\BThreeDsixtyfourDensityWins}{23}
\newcommand{\BThreeDsixtyfourNVNVFiniteCases}{23}

\newcommand{\BThreeDsixtyfourBoundedNVNVTime}{6.670336}
\newcommand{\BThreeDsixtyfourBoundedNVNVSpeedup}{1.186}

\newcommand{\BThreeThreeKStandardTflops}{3116}

\newcommand{\BThreeThreeKPeakSeq}{9472}

\newcommand{\BThreeThreeKPeakTflops}{3159}

\begin{document}

\maketitle
\begin{abstract}
Blackwell's 4-bit floating-point (FP4) tensor cores do not automatically make
attention faster because softmax conversion and on-chip dependencies dominate
once its matrix products shrink.  We address this with \emph{Direct-P} for
noncausal inference and a causal path that passes the forward quantization
directly into backward.
Direct-P maps scores directly to FP4 probabilities and reaches up to
2.13$\times$ the bfloat16 (BF16) forward throughput on an NVIDIA GB200.
The causal path reconstructs probabilities from saved quantized queries and
keys and uses 8-bit floating-point (FP8) gradient operands, accelerating a
complete single-GPU 8-billion-parameter update by up to 1.14$\times$.
Matched distributed training retains FP8 probabilities and values; every
tested MXFP4 probability/value training trajectory diverges.
\end{abstract}

\section{Introduction}

Attention contains two matrix products separated by softmax.  NVIDIA
Blackwell executes 4-bit floating-point (FP4) matrix multiplication much
faster than bfloat16 (BF16), but FP4 does not accelerate the work between
those products.  Softmax must still reduce each score tile, evaluate
exponentials, construct a scaled probability tile, and make that tile ready
for the second product.  This middle stage becomes the bottleneck as the
matrix products get faster.

Attention combines a query matrix $Q$, key matrix $K$, and value matrix $V$:
\begin{equation}
  S=QK^\mathsf{T}/\sqrt d,\qquad
  P=\softmax(S),\qquad
  O=PV.
  \label{eq:attention-intro}
\end{equation}
We call the two matrix products QK and PV.  Here $S$ contains similarity
scores, $P$ contains the resulting probabilities, and $O$ is the output.
FlashAttention evaluates these equations in tiles so that the quadratic score
and probability matrices need not be stored in high-bandwidth memory (HBM)
\cite{dao2022flashattention}.  Later versions improved how work is divided and
overlapped \cite{dao2023flashattention2,shah2024flashattention3}.
FlashAttention-4 (FA4) adapts this tiled algorithm to NVIDIA Blackwell's
asynchronous matrix hardware \cite{zadouri2026flashattention4}.

We study forward inference and causal training separately.  In the forward
pass, we ask whether all four operands inside attention---$Q$, $K$, $P$, and
$V$---can use FP4 and still outperform BF16 FA4.  We build on HAO AI Lab's
two-query Blackwell schedule and replace its probability path
\cite{hao2026fp4flashattention,zhang2026attnqat}.  Our method, \emph{Direct-P},
maps normalized scores directly to MXFP4 probability codes and normalizes the
same rounded values that the PV product consumes.  In this paper, ``full FP4''
refers only to these four attention operands; it does not mean that every
transformer operation uses FP4.

In training, we ask whether causal backward can reuse the low-precision state
created by the forward pass.  Forward passes the quantized $Q/K$ payload,
its scales, and the softmax normalizer directly to backward.  The projection
and gradient epilogues also publish the row- or column-oriented FP8 views
needed by each gradient matrix product.  Backward reconstructs the
probabilities from this state instead of creating a separate BF16 score path.
Learned projections remain a separate precision boundary, so this is not
pure-FP4 end-to-end training.

The paper makes three contributions:
\begin{enumerate}
  \item Direct-P reaches up to 2.13$\times$ the BF16 forward throughput on
  favorable Blackwell shapes; matched fixed-input evaluations measure its
  error against an FP8 probability path.
  \item Passing the forward quantization into backward accelerates
  projection-inclusive attention and a
  single-GPU 8-billion-parameter update.  Distributed diagnostics select FP8
  P/V because every tested MXFP4 P/V trajectory diverges.
  \item Hardware profiles identify tensor-memory ownership as the main limit on
  overlap and motivate another allocatable score destination with compatible
  issue semantics.
\end{enumerate}

The public implementation and reproduction materials are linked in
Appendix~\ref{app:reproduction}.

Table~\ref{tab:reader-notation} defines the notation and hardware terms used
throughout the paper.  E$x$M$y$ names a floating-point format with $x$
exponent bits and $y$ explicit fraction bits; for example, E4M3 is an 8-bit
format and E2M1 is the payload used by the FP4 formats studied here.

\begin{table}[htbp]
\centering
\caption{Reader's guide to the main notation and Blackwell hardware terms.}
\label{tab:reader-notation}
\small
\begin{tabularx}{\textwidth}{p{.21\textwidth}X}
\toprule
Term & Meaning \\
\midrule
$B,S,H,H_q,H_{kv},D$ & Batch size, sequence length, head count, query-head
count, key/value-head count, and per-head dimension. \\
Shape shorthand & Compact labels append each value: B1/S4096/H24/D128 means
batch 1, sequence length 4096, 24 heads, and head dimension 128. \\
FP32, BF16, FP8, FP4 & 32-, 16-, 8-, and 4-bit floating-point families.
Smaller formats increase matrix throughput but need explicit scaling. \\
NVFP4 and MXFP4 & The two block-scaled FP4 families used here.  NVFP4 uses
fine-grained data-dependent scales; MXFP4 shares one power-of-two scale across
each 32-value block. \\
SM and CTA & A graphics processing unit (GPU) contains streaming
multiprocessors (SMs).  A cooperative thread array (CTA) is a CUDA thread
block scheduled on an SM. \\
TMEM & Tensor memory: Blackwell's on-chip accumulator scratchpad for
asynchronous matrix operations. \\
TMA and MMA & The Tensor Memory Accelerator (TMA) moves tiles; a matrix
multiply--accumulate (MMA) instruction performs the tensor-core product. \\
\bottomrule
\end{tabularx}
\end{table}

Sections~\ref{sec:hao-background}--\ref{sec:direct-p} develop the forward
argument in order: inherited schedule, remaining problems, and proposed
fixes.  Section~\ref{sec:causal-training} keeps isolated backward,
projection-inclusive attention, complete model updates, and distributed
training trajectories as separate measurement boundaries.
Section~\ref{sec:current-headroom} then relates the measured bottlenecks to
hardware and closes with the evidence boundaries and conclusions.

\section{Previous Work}
\label{sec:hao-background}

Direct-P starts from HAO AI Lab's FP4 FA4 implementation
\cite{hao2026fp4flashattention}.  HAO already provides a two-query pipeline
that assigns data movement, matrix products, softmax, and output correction to
specialized warps.  We keep that outer schedule.  Our forward contribution is
the narrower stage that turns a completed score fragment into the scaled
probability operand consumed by the value product.  This section first
explains the tiled algorithm, then states exactly which scheduling and storage
decisions we inherit.

\subsection{Why FlashAttention uses an online softmax}

For one query row, exact attention is
\begin{equation}
  O_i=\frac{\sum_j e^{z_{ij}}V_j}{\sum_j e^{z_{ij}}},
  \qquad z_{ij}=Q_iK_j^\mathsf{T}/\sqrt d.
  \label{eq:exact-attention}
\end{equation}
FlashAttention visits the keys in tiles.  After a tile with maximum
$\widetilde m_i$ arrives, it updates a running maximum $m_i$, denominator
$\ell_i$, and output $o_i$:
\begin{align}
  m_i' &= \max(m_i,\widetilde m_i), \\
  \ell_i' &= e^{m_i-m_i'}\ell_i+
    \sum_{j\in B}e^{z_{ij}-m_i'}, \\
  o_i' &= e^{m_i-m_i'}o_i+
    \sum_{j\in B}e^{z_{ij}-m_i'}V_j.
  \label{eq:online-recurrence}
\end{align}
This recurrence avoids storing the full score and probability matrices in
HBM.  Its cost is a strict dependency: a score tile must be reduced and turned
into probabilities before those probabilities can be multiplied by $V$.

On data-center Blackwell GPUs, asynchronous \tcgen{} operations accumulate
32-bit floating-point (FP32) score and output tiles in tensor memory (TMEM).
The Tensor Memory Accelerator (TMA) supplies shared-memory operands.  Separate
warps issue matrix operations, compute softmax, and correct the online output
\cite{nvidia2026ptx,zadouri2026flashattention4}.

\subsection{Implementation baseline}

HAO's implementation supports block-scaled FP4 QK with BF16 or FP8 PV, as
well as a stabilized path that uses NVFP4 throughout attention.  Its
implementation and Attn-QAT, an attention quantization-aware-training study,
identify online P quantization and scale movement as central costs
\cite{hao2026fp4flashattention,zhang2026attnqat}.

SageAttention3 contributes related numerical ideas, including Q/K smoothing,
two-level P scaling, and reuse of the block maximum during online softmax
\cite{zhang2025sageattention3}.  It targets a consumer Blackwell interface
that differs from the asynchronous tensor-core and tensor-memory interface on
the data-center Blackwell GPUs evaluated here \cite{nvidia2026ptx}.  We use
those numerical ideas as context, but inherit the execution schedule from HAO
and FA4.

\subsection{Two-query execution model}

For our purposes, HAO's key contribution is a complete ownership plan for the
Blackwell pipeline \cite{hao2026fp4flashattention}.  At head dimension 128
(D128), one 16-warp cooperative thread array (CTA) advances two 128-row query
tiles (M128), called stage 0 and stage 1.  A load warp keeps K and V ahead in a
TMA-fed shared-memory ring.  One warp issues ordered QK and PV matrix
multiply--accumulate (MMA) operations.  Two softmax warpgroups (WGs), each
made of four warps, process the query stages in ping-pong fashion.  Separate
warps correct the online state and store the final output.

Small hardware barriers act as event counters rather than stopping the whole
CTA.  They announce ``score ready'', ``first P half ready'', ``tail P ready'',
and ``output released''.  Algorithm~\ref{alg:hao-pipeline} states the retained
ownership protocol.  Operations for $q=0$ and $q=1$ interleave whenever their
dependencies allow; the table gives the required order for one score bank,
not a serialized whole-CTA loop.

N32 denotes a 32-column score or probability fragment.  K$x$ denotes an
$x$-element matrix-product inner dimension, so K64 consumes two adjacent N32
fragments.

\begin{algorithm}[htbp]
\caption{Two-query score-to-PV ownership protocol inherited from HAO}
\label{alg:hao-pipeline}
\small
\begin{tabularx}{\textwidth}{@{}r p{.18\textwidth} X@{}}
\toprule
& Owner & Event for query stage $q\in\{0,1\}$ \\
\midrule
1 & Load warp & Prefetch the next K/V tile into the shared-memory ring. \\
2 & MMA warp & Wait until score bank $S_q$ is free; issue QK into $S_q$ and publish \emph{score ready}. \\
3 & Softmax WG $q$ & Read N32 quarters Q0 and Q1, create packed P and scales in retired score addresses, then publish the first legal K64 operand. \\
4 & MMA warp & Observe the first-half event and issue asynchronous $PV_{q,0}$ into permanent output bank $O_q$. \\
5 & Softmax WG $q$ & Process Q2 and Q3 and publish the tail K64 operand. \\
6 & MMA warp & Issue $PV_{q,1}$; meanwhile issue legal work for stage $1-q$. \\
7 & Correction WG & Update online state; the epilogue normalizes and stores when the key loop ends. \\
8 & MMA warp & After tail PV has consumed the overlay, publish $S_q$ as free for the next QK tile. \\
\bottomrule
\end{tabularx}
\end{algorithm}

The two-query topology matters at high head count because one CTA performs two
query jobs while reusing the same K/V stream.  A one-query topology creates
twice as many independent CTAs and can add a second scheduling wave after all
streaming multiprocessors are occupied.  A third query stage would provide
more look-ahead, but the TMEM layout leaves no bank for it.

\subsection{Why the D128 layout cannot look farther ahead}

SM100 and the tested SM103 path expose 512 logical TMEM columns. A D128 FP32
score tile or output accumulator uses 128 columns. Two score banks and two
output banks therefore consume the entire allocation.

\begin{figure}[htbp]
\centering
\begin{tikzpicture}[x=0.044\textwidth,y=0.72cm,font=\small]
  \draw[thick] (0,0) rectangle (20,1);
  \fill[scoreblue!22] (0,0) rectangle (5,1);
  \fill[scoreblue!36] (5,0) rectangle (10,1);
  \fill[outputgreen!26] (10,0) rectangle (15,1);
  \fill[outputgreen!40] (15,0) rectangle (20,1);
  \draw (5,0) -- (5,1);
  \draw (10,0) -- (10,1);
  \draw (15,0) -- (15,1);
  \node at (2.5,.55) {$S_0$: score stage 0};
  \node at (7.5,.55) {$S_1$: score stage 1};
  \node at (12.5,.55) {$O_0$: output stage 0};
  \node at (17.5,.55) {$O_1$: output stage 1};
  \node[below] at (0,0) {0};
  \node[below] at (5,0) {128};
  \node[below] at (10,0) {256};
  \node[below] at (15,0) {384};
  \node[below] at (20,0) {512};
  \node[align=center,text=inkgray,font=\scriptsize,text width=.40\textwidth]
    at (5,-1.05)
    {temporary: QK score, then packed $P$ and scale pages};
  \node[align=center,text=inkgray,font=\scriptsize,text width=.40\textwidth]
    at (15,-1.05)
    {long lived: FP32 online output accumulators};
\end{tikzpicture}
\caption{D128 TMEM layout inherited from HAO. P and its instruction-facing
scales reuse score storage only after that score fragment is retired. There
is no unowned 128-column bank.}
\label{fig:tmem-layout}
\end{figure}
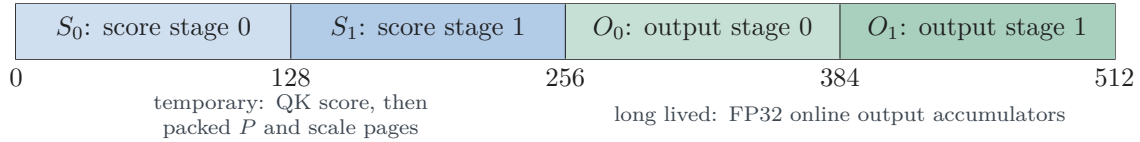

HAO reuses each score bank as
\begin{equation}
  \text{QK score}\rightarrow\text{read N32 fragment}
  \rightarrow\text{P/scale overlay}\rightarrow\text{PV consume}
  \rightarrow\text{free}.
  \label{eq:score-overlay}
\end{equation}
Writing the next QK tile too early destroys P while PV still needs it.
Waiting leaves the matrix issuer idle.  The scheduling problem is therefore a
storage-ownership handoff, not merely a shortage of barrier signals.  A
barrier can report that storage is free; it cannot create another destination.

\subsection{Why probability tiles are published in halves}

QK still writes one 128-row by 128-column (M128$\times$N128) FP32 score tile.
``Quartering'' means
that a softmax warpgroup reads and transforms four N32 fragments,
\begin{equation}
  S=[Q0\mid Q1\mid Q2\mid Q3],\qquad
  Qk\in\mathbb R^{128\times32},
\end{equation}
not that QK becomes four smaller matrix products. The score allocation remains
128 columns.

The scaled-FP4 PV instruction used by the retained HAO path consumes K64
rather than K32. Two adjacent quarters must therefore be complete before
useful matrix work can issue:
\begin{equation}
  (Q0,Q1)\rightarrow PV_{K64}^{(0)},\qquad
  (Q2,Q3)\rightarrow PV_{K64}^{(1)}.
  \label{eq:k64-pairs}
\end{equation}
Quartering reduces the live register fragment and lets P overwrite score
addresses that are no longer needed.  It does not reduce the score bank's
TMEM allocation.  Its performance value is the event after Q1: first-half PV
can begin while Q2/Q3 and the other query stage provide independent work.

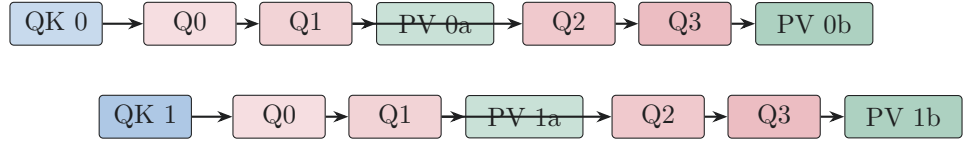
\begin{figure}[htbp]
\centering
\begin{tikzpicture}[
  x=1.18cm,y=.82cm,font=\small,
  box/.style={draw,rounded corners=1.5pt,minimum width=1.22cm,
              minimum height=.56cm,align=center},
  dep/.style={-{Stealth[length=2mm]},thick}]
  \node[box,fill=scoreblue!26] (s0) at (0,1.5) {QK 0};
  \node[box,fill=probred!17] (a0) at (1.5,1.5) {Q0};
  \node[box,fill=probred!23] (a1) at (2.8,1.5) {Q1};
  \node[box,fill=outputgreen!25,minimum width=1.55cm] (p0) at (4.25,1.5) {PV 0a};
  \node[box,fill=probred!28] (a2) at (5.75,1.5) {Q2};
  \node[box,fill=probred!34] (a3) at (7.05,1.5) {Q3};
  \node[box,fill=outputgreen!38,minimum width=1.55cm] (p1) at (8.5,1.5) {PV 0b};

  \node[box,fill=scoreblue!38] (s1) at (1.0,0) {QK 1};
  \node[box,fill=probred!17] (b0) at (2.5,0) {Q0};
  \node[box,fill=probred!23] (b1) at (3.8,0) {Q1};
  \node[box,fill=outputgreen!25,minimum width=1.55cm] (p2) at (5.25,0) {PV 1a};
  \node[box,fill=probred!28] (b2) at (6.75,0) {Q2};
  \node[box,fill=probred!34] (b3) at (8.05,0) {Q3};
  \node[box,fill=outputgreen!38,minimum width=1.55cm] (p3) at (9.5,0) {PV 1b};

  \draw[dep] (s0)--(a0); \draw[dep] (a0)--(a1);
  \draw[dep] (a1)--(p0); \draw[dep] (a1)--(a2);
  \draw[dep] (a2)--(a3); \draw[dep] (a3)--(p1);
  \draw[dep] (s1)--(b0); \draw[dep] (b0)--(b1);
  \draw[dep] (b1)--(p2); \draw[dep] (b1)--(b2);
  \draw[dep] (b2)--(b3); \draw[dep] (b3)--(p3);
\end{tikzpicture}
\caption{Dependency sketch for the two query stages. Box width is not
measured time. Each first PV half waits for two N32 producer quarters; work
from the other stage covers part of that delay.}
\label{fig:quarter-pipeline}
\end{figure}

We inherit this outer pipeline, tensor-memory lifecycle, and publication
protocol unchanged.  They also expose the remaining problem: PV cannot begin
until softmax has constructed, scaled, packed, and published two adjacent
probability fragments.  The next section separates this timing constraint
from the numerical range constraint.

\section{What Prevents a Full-FP4 Speedup?}
\label{sec:forward-problems}

HAO's schedule already overlaps QK, softmax, and PV whenever their
dependencies allow.  Replacing its FP8 probability--value product with FP4
leaves two linked problems.  The first is timing: constructing P delays the
first useful PV instruction.  The second is numerical: softmax probabilities
span a range that is awkward for a 4-bit payload and its block scale.

\subsection{Problem 1: P lies on the critical path}

In tiled attention, P is produced inside the kernel rather than loaded as an
input.  For full FP4 attention, the first useful PV command therefore waits
for
\begin{equation}
  S\rightarrow\text{maximum}\rightarrow\text{scale}
  \rightarrow\text{probability}\rightarrow\text{E2M1 pack}
  \rightarrow\text{publish}.
  \label{eq:serial-p-path}
\end{equation}
Work removed after publication may not change latency. Work removed before
the first legal P tile can.

FP4 accelerates QK and PV, but not score reduction, exponentiation,
synchronization, or scale publication.  It also requires P to be converted
between the two products.  Removing work before the first legal P tile can
shorten the critical path; removing work after that publication may not
change latency \cite{nvidia2026ptx,zadouri2026flashattention4}.

\subsection{Problem 2: P needs both range and cheap scales}

NVFP4 and MXFP4 both use signed E2M1 payloads.  Ignoring sign, their
nonnegative magnitudes are
\begin{equation}
  \mathcal F_{\mathrm{E2M1}}=
  \left\{0,\tfrac12,1,\tfrac32,2,3,4,6\right\}.
\end{equation}
The formats differ in how they scale these payloads:
\begin{table}[H]
\centering
\caption{Block-scaled FP4 formats used in this work.}
\label{tab:fp4-formats}
\small
\begin{tabularx}{\textwidth}{lcccX}
\toprule
Format & Block & Encoded scale & Main strength & Role in this work \\
\midrule
NVFP4 & 16 & E4M3, optional tensor scale & fine local placement & signed Q and K \\
MXFP4 & 32 & E8M0 amplitude, power of two & wide exponent range & P and V operands \\
\bottomrule
\end{tabularx}
\end{table}

For an NVFP4 P block with maximum $a_B$ and global encode factor $G$, the
decoded block scale is approximately
\begin{equation}
  \widehat s_{\mathrm{NV}}=
  \frac{Q_{\mathrm{E4M3}}(G a_B/6)}{G}.
  \label{eq:nv-scale}
\end{equation}
If the E4M3 scale rounds to zero, the entire block disappears. A row-level
stabilizer fixes the range, but its scale and inverse correction add state to
the online path. MXFP4 instead stores a power-of-two block amplitude near
$a_B$,
\begin{equation}
  \alpha_B\in
  \left\{2^{\lfloor\log_2 a_B\rfloor},
          2^{\lceil\log_2 a_B\rceil}\right\},
  \qquad \delta_B=\frac{\alpha_B}{6},
  \label{eq:mx-scale}
\end{equation}
where $\alpha_B$ is the E8M0 value written to the scale page and $\delta_B$
is the effective reconstruction step.  Since the largest E2M1 code is 6, our
convention reconstructs one operand as $\alpha_B q/6$ and corrects an MXFP4
P/V product by $1/36$.  The amplitude folds naturally into a base-two score
transform and matches one 32-column producer fragment (N32). The choice is a
latency--range trade-off, not a claim that E8M0 is intrinsically more accurate
than E4M3.

Table~\ref{tab:p-range} isolates numerical range from kernel scheduling. It
forms exact Gaussian softmax probabilities, quantizes matched N32 blocks,
renormalizes the represented P, and multiplies by the same FP32 V. Stabilized
NVFP4 is the high-fidelity control; unscaled NVFP4 exposes underflow.

\begin{table}[htbp]
\centering
\caption{Probability-format range at D128. ``Zero scales'' is the fraction
of blocks whose scale encodes as zero; ``lost mass'' is exact probability
mass mapped to zero. This is a numerical diagnostic, not a kernel timing.}
\label{tab:p-range}
\scriptsize
\begin{adjustbox}{max width=\textwidth}
\begin{tabular}{llrrrrrr}
\toprule
$S$ & Format & Zero scales & Zero payload & Lost mass & $P$ rel.-$L_2$
& $PV$ cosine & $PV$ rel.-$L_2$ \\
\midrule
1024 & NVFP4 G{=}1 & 0.706299 & 0.831829 & 0.683773 & 1.715156 & 0.700819 & 1.685442 \\
1024 & NVFP4 G{=}448 & 0.000000 & 0.155418 & 0.029314 & 0.114408 & 0.993854 & 0.113687 \\
1024 & MXFP4 & 0.000000 & 0.188683 & 0.040178 & 0.146380 & 0.989846 & 0.144944 \\
4096 & NVFP4 G{=}1 & 0.996063 & 0.999438 & 0.994332 & 13.815642 & 0.217328 & 13.876892 \\
4096 & NVFP4 G{=}448 & 0.000000 & 0.156454 & 0.029760 & 0.114091 & 0.993747 & 0.114582 \\
4096 & MXFP4 & 0.000000 & 0.189755 & 0.040494 & 0.147209 & 0.989338 & 0.148716 \\
8192 & NVFP4 G{=}1 & 0.999634 & 0.999977 & 0.999284 & 12.467307 & 0.096774 & 12.389232 \\
8192 & NVFP4 G{=}448 & 0.000000 & 0.160613 & 0.030691 & 0.114725 & 0.993841 & 0.113660 \\
8192 & MXFP4 & 0.000000 & 0.194189 & 0.041630 & 0.147318 & 0.989660 & 0.146237 \\
\bottomrule
\end{tabular}
\end{adjustbox}
\end{table}

The alternatives therefore impose different costs.  Stabilized NVFP4 offers
the best FP4 fidelity but needs a row-level range correction.  FP8 avoids the
4-bit range problem but gives up the FP4 PV instruction.  MXFP4 provides a
power-of-two scale at the same N32 granularity as the producer fragment, but
places probabilities more coarsely.  Direct-P chooses MXFP4 and redesigns how
that operand is produced and normalized.

\section{Direct-P: Fixing the FP4 Probability Path}
\label{sec:direct-p}

\subsection{Design overview}

The previous section identified one timing problem and one numerical problem.
Direct-P addresses both within a narrow boundary: we preserve HAO's outer
schedule and change only the interval from a ready FP32 score fragment to a
legal FP4 probability operand for PV.  Table~\ref{tab:hao-vs-ours} separates
the inherited machinery from our changes.

\begin{table}[htbp]
\centering
\caption{Inherited structure and changes made in this work.}
\label{tab:hao-vs-ours}
\small
\begin{tabularx}{\textwidth}{p{.24\textwidth}Y Y}
\toprule
Component & Inherited from HAO & This work \\
\midrule
CTA and TMEM & Two query stages, two score banks, two FP32 output banks, one ordered MMA issuer. & Retained. \\
P publication & N32 producer quarters, first-half and tail barriers, score/P overlay. & Retained; less work before each event. \\
Formats & Full-FP4 comparator: NVFP4 Q/K/P/V with stabilized P. & NVFP4 Q/K, MXFP4 P/V. \\
P arithmetic & Exponential evaluation followed by block-scale quantization. & Direct log-score-to-E2M1 map with selective hardware exponentials. \\
Normalization & Denominator accumulated from floating exponential values. & Denominator accumulated from the represented P consumed by PV. \\
\bottomrule
\end{tabularx}
\end{table}

Within the noncausal forward benchmark boundary, Q, K, and V are prequantized.
QK uses NVFP4 with adjacent K64 Q/K scales folded offline by a
mean-squared-error (MSE) rule. For P/V, each N32 probability block uses one
MXFP4 E8M0 scale. P payload and scale pages overwrite retired score addresses,
exactly as required by the lifecycle in Algorithm~\ref{alg:hao-pipeline}.

Direct-P makes three linked choices.  First, it maps normalized scores
directly to the E2M1 payload consumed by PV, shortening the critical path.
Second, it computes the normalizer from those same rounded payloads, so the
numerator and denominator describe one approximate operator.  Third, it adds
a range guard only for model layers with extreme logits.  The first choice
addresses timing; the other two keep the cheaper representation numerically
well defined.

\subsection{Fix 1: map scores directly to E2M1 codes}

The standard route computes a relatively accurate exponential and then rounds
it to one of eight nonnegative E2M1 magnitudes.  That intermediate precision does
not reach PV.  Direct-P instead treats probability construction as a code
classification problem: determine which E2M1 bin each normalized score should
occupy.

Let $m_i$ be a row reference and
$a_B=\max_{j\in B}\exp(z_{ij}-m_i)$. For a nonzero N32 block, let $u_B$ be
the biased E8M0 byte, $e_B=u_B-127$ its exponent, and
$\alpha_B=2^{e_B}$ its stored amplitude. The effective E2M1 reconstruction
step is $\delta_B=\alpha_B/6$, so Direct-P represents a probability as
\begin{equation}
  \widetilde p_{ij}=\delta_B q_{ij}=\alpha_B q_{ij}/6,
  \qquad q_{ij}\in\mathcal F_{\mathrm{E2M1}}.
  \label{eq:represented-probability}
\end{equation}
Thus $u_B$ encodes $\alpha_B$, not $\delta_B$.  The scaled matrix instruction
uses $\alpha_Bq_{ij}$; its output epilogue applies the fixed $1/36$ correction
for MXFP4 P and V.
The ideal converter input is
\begin{align}
  x_{ij} &= (z_{ij}-m_i)\log_2 e-e_B+\log_2 6, \\
  q_{ij} &= Q_{\mathrm{E2M1}}(2^{x_{ij}}).
  \label{eq:ideal-code-path}
\end{align}
Only the code $q_{ij}$ reaches PV. Computing an accurate FP32 $2^x$ and then
discarding almost all of its precision is unnecessary.

Positive E2M1 changes value at seven rounding boundaries,
\begin{equation}
  \left\{\tfrac14,\tfrac34,\tfrac54,\tfrac74,
  \tfrac52,\tfrac72,5\right\}.
  \label{eq:e2m1-boundaries}
\end{equation}
We therefore fit an affine classifier in value space,
\begin{equation}
  \widehat u(x)=\max(0,Ax+B),\qquad
  \widehat q(x)=Q_{\mathrm{E2M1}}(\widehat u(x)),
  \label{eq:affine-code-map}
\end{equation}
to maximize E2M1 code agreement rather than real-valued exponential
accuracy. The score transform, $e_B$, and $\log_2 6$ term are folded into the
packed fused-multiply--add (FMA) coefficients. A packed two-lane FMA
instruction (\code{FFMA2}) handles two scores, and a packed floating-point
conversion instruction (\code{F2FP}) emits their E2M1 payloads. A selected
pair may instead use native EX2, Blackwell's base-two exponential instruction,
when that improves the machine schedule.

\begin{algorithm}[H]
\caption{Direct MXFP4 probability production for one N32 score fragment}
\label{alg:direct-p-pack}
\small
\begin{tabularx}{\textwidth}{@{}rX@{}}
\toprule
1 & Load 32 scores per row from the completed TMEM score tile and reduce the block maximum. \\
2 & Select power-of-two amplitude $\alpha_B=2^{e_B}$ and biased E8M0 byte $u_B=e_B+127$ from that maximum and the row reference $m_i$; $u_B=0$ denotes a zero block. \\
3 & For each packed score pair, form $x=(z-m_i)\log_2 e-e_B+\log_2 6$. \\
4 & Evaluate either $\max(0,Ax+B)$ or selected native $2^x$, then use packed conversion to emit E2M1 codes $q$. \\
5 & Pack four payload words and decode their small represented sum $c_B=\sum_{j\in B}q_{ij}$. \\
6 & Accumulate denominator contribution $d_{iB}=\alpha_B(c_B/6)$, using the same amplitude and codes as PV. \\
7 & Overlay P payload and the swizzled scale page onto retired score addresses. Publish after two adjacent N32 fragments form one legal K64 operand. \\
\bottomrule
\end{tabularx}
\end{algorithm}

The generic \code{fast} fit uses $A=1.50,B=1.20$. Wan activation evaluations
select the equal-cost pair $A=1.60,B=0.95$. These constants move the seven
decision boundaries,
\begin{equation}
  x_j=(t_j-B)/A,
  \label{eq:affine-log-boundaries}
\end{equation}
where $t_j$ is an E2M1 boundary. They are quantizer-calibration parameters,
not changes to exact softmax. Layer-wise maps were tested but not retained:
isolated substitutions on a BF16 teacher trajectory did not predict the
composed all-FP4 trajectory (Appendix~\ref{app:joint-affine}).

Integer threshold trees, lookup tables (LUTs), and custom nibble packing
generated more NVIDIA machine code (SASS) than packed FMA followed by
Blackwell's native converter. Quadratic and cubic
fits improved real-function error but placed extra dependent FMA operations
directly before publication.

\subsection{Fix 2: normalize the represented probability}

The numerator uses rounded FP4 probabilities, so an independently approximated
floating-point denominator would describe a different operator.  Direct-P
instead builds the denominator from the exact codes and block scales consumed
by PV.

For block $B$, the numerator and denominator actually consumed by the
approximate operator are
\begin{align}
  \widetilde N_{iB} &= \frac{\alpha_B}{6}
    \sum_{j\in B}q_{ij}\widehat V_j, \\
  \widetilde L_{iB} &= \frac{\alpha_B}{6}
    \sum_{j\in B}q_{ij}.
  \label{eq:represented-contributions}
\end{align}
where $\widehat V_j$ is the reconstructed MXFP4 value, including its own
$1/6$ correction.  The output is
\begin{equation}
  \widetilde O_i=
  \frac{\sum_B\widetilde N_{iB}}{\sum_B\widetilde L_{iB}}.
  \label{eq:represented-output}
\end{equation}
This avoids normalizing an E2M1 numerator with an unrelated approximate FP32
exponential sum. Four packed payload words are reduced with byte permutations,
carry-free byte addition, and one four-way integer dot-product-accumulate
instruction (\code{DP4A}). Streaming each word or keeping
pairwise partials is algebraically equivalent but slower because it extends
live ranges and changes instruction interleaving.

FA4 routes some exponential work from special-function units (SFUs) to
ordinary arithmetic pipelines
\cite{zadouri2026flashattention4}. We apply the same balancing principle to
the code map. GB200 D128 \code{fast} is all-affine; \code{accurate} uses
native EX2 on roughly one quarter of pair positions. B300 has stronger SFUs,
but broad EX2 routing still loses. Its retained policy uses native EX2 only in
the quarter and shape regimes where the measured dependency schedule benefits.

\subsection{Fix 3: guard only extreme model logits}

The fast shiftless path is finite on the synthetic grid and most model
layers, but a few late Wan layers produce BF16 logits above 500 and sometimes
1000. Scanning or reloading every score would remove the speed advantage. We
instead route only those layers through Algorithm~\ref{alg:sampled-guard}.
Before launch, 128 globally distributed K/V rows are moved together into the
first physical key tile. This permutation leaves exact attention unchanged.

\begin{algorithm}[H]
\caption{Sampled QK guard and underflow-safe represented denominator}
\label{alg:sampled-guard}
\small
\begin{tabularx}{\textwidth}{@{}rX@{}}
\toprule
1 & In the first key tile, reduce all four N32 quarters and set sampled anchor $a_i=\max_{j\in\mathcal A}z_{ij}$, where $|\mathcal A|=128$. \\
2 & Choose compile-time margin $M$ and stored-scale shift $L$ with $M+L\le126$; use row reference $m_i=a_i+M\ln2$. \\
3 & Follow Algorithm~\ref{alg:direct-p-pack}, but floor the working byte before forming $x$ and $q$: $\bar u_B=\max(u_B,L+1)$ and $\bar e_B=\bar u_B-127$. Publish $u'_B=\bar u_B-L$. \\
4 & Let $\alpha'_B=2^{u'_B-127}$ and $c_B=\sum q_{ij}$. Evaluate the represented denominator as $\alpha'_B(c_B/6)$ in that order. \\
5 & Never evaluate $(\alpha'_B/6)c_B$: at $u'_B=1$, the first product is subnormal and may flush a valid block to zero. \\
\bottomrule
\end{tabularx}
\end{algorithm}

The working floor preserves $\bar u_B-u'_B=L$ across the byte range; flooring
only the published byte would amplify blocks with $u_B\le L$.  The common
$2^{-L}$ factor cancels between numerator and denominator, and a saved log
normalizer restores the corresponding exponent shift.

The 1.3-billion-parameter Wan model uses $(M,L)=(110,16)$ and the
14-billion-parameter model uses $(112,14)$, so both consume the
full 126-step budget. The bound is enforced at compile time and translated
E8M0 scales are floored at code 1. The reordered evaluation uses the same two
multiplies as the failing expression. On the original layer-39 failure, the
sampled anchor already equalled the exact row maximum; the failure was solely
the subnormal intermediate. This correction fixes it without a second scan,
new barrier, or stable-softmax fallback.

\subsection{Two operating points}

\begin{table}[htbp]
\centering
\caption{Retained policies. Both use NVFP4 QK, MXFP4 P/V, K64 PV issue, and
the two-query HAO layout.}
\label{tab:policy-definition}
\small
\begin{tabularx}{\textwidth}{lccccX}
\toprule
Policy & K/V stages & Anchor & Native EX2 & Denominator & Goal \\
\midrule
\code{fast} & 12 & none by default & 0 on GB200 & producer & minimum latency \\
\code{accurate} & 13 & 32 fixed rows & about 25\% & correction WG & higher fidelity \\
\bottomrule
\end{tabularx}
\end{table}

The Wan bundle can add the 128-row routed guard to either base policy. Anchor
permutations and folded Q/K scale preparation are outside the timed attention
kernel and should be fused into an upstream layout or quantization step in a
deployment.

\FloatBarrier

\section{Experimental Setup}

We keep the measurement boundaries separate.  Kernel speed does not include
projection or optimizer work, and a short fixed-token update does not establish
training quality.  Table~\ref{tab:measurement-boundaries} states what each
comparison supports.

\begin{table}[htbp]
\centering
\caption{Measurement boundaries used in the paper.}
\label{tab:measurement-boundaries}
\small
\begin{tabularx}{\textwidth}{p{.25\textwidth}p{.40\textwidth}X}
\toprule
Boundary & Included work & Supported conclusion \\
\midrule
Noncausal forward kernel & QK, online softmax, P publication, PV, and output
epilogue & Forward latency and output error. \\
Causal backward kernel & Probability reconstruction and attention gradients;
operands are prepared before timing & Backward latency and gradient
correctness. \\
Projection-inclusive attention & QKV projection, rotary embedding, operand
publication, attention, output projection, and gradients & Whether the
attention gain survives its immediate producers and consumers. \\
Single-GPU 8B update & Complete model forward, loss, backward, and optimizer &
End-to-end step time at a fixed local batch. \\
Distributed trajectory & Data loading, communication, checkpointing, and
validation & Observed training stability, loss, and sustained throughput. \\
\bottomrule
\end{tabularx}
\end{table}

\subsection{Metrics and comparison rules}

Each principal row stores latency and numerical error from one deterministic
case. Intentionally incomplete kernels and older runs with different seeds or
timing windows are used only as diagnostics. For approximate output
$\widehat O$ and BF16 reference $O$, we report
\begin{align}
  \operatorname{cos}(\widehat O,O)
    &=\frac{\langle\widehat O,O\rangle}
      {\lVert\widehat O\rVert_2\lVert O\rVert_2}, \\
  \operatorname{rel}L_2
    &=\frac{\lVert\widehat O-O\rVert_2}{\lVert O\rVert_2}, \\
  \operatorname{RMSE}
    &=\sqrt{n^{-1}\sum_r(\widehat O_r-O_r)^2}.
\end{align}
Cosine can hide magnitude error and root-mean-square error (RMSE) depends on
output scale, so relative-$L_2$ is reported alongside both.

\subsection{Operator benchmark suites}

The primary GB200 suite reproduces the D128 shape grid in HAO's FP4 FA4
README:
\begin{equation}
\begin{split}
 &(B,S,H)\in\{(1,256,16),(1,1024,16),
 (4,4096,16),\\
 &(1,32768,16),(4,4096,32),(1,4096,12),
 (1,32768,12),\\
 &(1,4096,24),(1,32768,24)\}.
\end{split}
\label{eq:hao-grid}
\end{equation}
Here $B$, $S$, and $H$ are batch size, sequence length, and query-head count;
$D_{QK}$ and $D_{VO}$ below are the inner dimensions of the two matrix
products.
We name routes by their QK/PV formats: NV/NV means NVFP4/NVFP4, NV/MX means
NVFP4/MXFP4, and NV/FP8 means NVFP4/FP8.
Every case is noncausal D128, uses HAO's
\code{create\_nvfp4\_attention\_tensors} factory and seed 20260814, and is
timed with 300 ms warmup and a 3000 ms median window. ThunderKittens (TK)
\code{fast}, TK \code{accurate}, native HAO NV/NV, and HAO's generated
CuTe domain-specific-language (DSL) BF16 kernel receive the same
prepared Q/K/V. The two TK binaries run in separate processes, but both use
the controls from one canonical manifest:
\begin{equation}
  \operatorname{speedup}(k)=t_{\mathrm{HAO\ BF16}}/t_k.
  \label{eq:canonical-speedup}
\end{equation}
Throughput follows HAO's matrix-operation convention,
\begin{equation}
  \mathrm{ops}=BH\,2S^2(D_{QK}+D_{VO}).
  \label{eq:hao-flops}
\end{equation}

A saturation suite adds S4096/H64 and S8192/H64 plus local HAO and TK NV/FP8
controls. Its six-order harness removes short-case provider-order bias. A
separate D64 matrix is reported in Appendix~\ref{app:nvnv-d64}; D64 uses a
different tile and ownership regime and is not part of the main comparison.

B300 D128 rows are the best stable records from the final SM103 tuning
campaign, not a same-binary hardware A/B. S4096 uses 300/3000 ms windows;
S6144 and S32768 use repeated windows; S8192 and the wave-aligned S9472 case
use five 2000-iteration windows and two correctness seeds. Published HAO B200
and GB300 values are labeled as cross-run context rather than local timings.

\subsection{Timed scope and operand contract}

Kernel time includes score loading, NVFP4 QK, online state, P approximation
and packing, MX scale publication, MXFP4 PV, correction, and the output
epilogue. It excludes dynamic Q/K/V quantization and optional K/V
permutations, matching HAO's attention-kernel scope rather than full layer
latency.

The fast path requires adjacent NVFP4 Q/K scale blocks folded for K64 access.
The fold is chosen offline by reconstruction MSE. A production QKV projection
should emit this layout directly. Pairing the binary with ordinary block-16
scales is an invalid operand contract even if its timing looks plausible.

The causal-training extension reports the learned-projection format separately
from the attention format.  The study crosses E4M3 and NVFP4 learned QKV/O
projections with FP8 and MXFP4 attention P/V.  The QKV projection epilogue
publishes the Q/K/V layouts while its output fragments are live, avoiding
separate quantization and transpose kernels.  The current throughput arm uses
NVFP4 learned projections; E4M3 is the projection-accuracy control.  Historical
tables retain their actual projection formats and cut-cross-entropy loss
boundary.  The retained route instead uses a dense language-model head and
ordinary cross entropy, with compilation applied only to the loss.  We do not
transfer timing or loss values between these recipe generations.

\subsection{Fixed-input model evaluations}

Six fixed-input evaluations replace every eligible attention call in a Vision
Transformer (ViT) and Bidirectional Encoder Representations from Transformers
(BERT): ViT classification at S256, S1024, and S4096; BERT masked-language
modeling (MLM) at S256 and S512; and Stanford Sentiment Treebank v2 (SST-2)
classification at S256. All providers receive the same weights, examples,
masks, prepared operands, and BF16 reference digest. We report attention-layer
error, final logits or hidden-state error, prediction agreement, and task
score. These are held-input inference comparisons, not training-stability
claims.

The Vision Transformer masked-autoencoder (ViT-MAE) experiment replaces all
12 encoder self-attention layers for 100 Common Objects in Context (COCO)
validation images with a fixed 75\% patch mask
\cite{he2022mae,lin2014coco}. Its D64 model attention is padded to the existing
S256/H16/D128 specialization and padded keys are masked. Paired masked-patch
peak signal-to-noise ratio (PSNR), MSE, reconstruction cosine, and
relative-$L_2$ are measured against the same BF16 decoder run.

Wan2.1 uses the native noncausal D128 topology at S7680: H12 across 30 layers
for the 1.3-billion-parameter model and H40 across 40 layers for the
14-billion-parameter model \cite{wanteam2025wan}. Every video self-attention
call is replaced; cross-attention and the rest of the model remain BF16. The
fast route uses the global $A=1.60,B=0.95$ map. The sampled guard in
Algorithm~\ref{alg:sampled-guard} is routed only to layers 27--29 in the
1.3-billion-parameter model and 33--34/38--39 in the 14-billion-parameter
model. One-, four-, and twenty-step outputs are paired against HAO CuTe-DSL
BF16 with identical prompt, seed, guidance, and initial latent. A second
14-billion-parameter prompt/seed checks the 20-step fix. Kernel timing uses
independently warmed five-second windows and weights guarded and unguarded
latencies by their layer counts.

Layer-wise affine calibration is kept as a negative control, not part of the
reported route. Isolated substitutions were scored on a BF16 teacher
trajectory, then rejected because the composed all-FP4 route changed sign
between calibration and held-out prompts. The full control is in
Appendix~\ref{app:joint-affine}.

\subsection{Reproducibility}

Generated tables are built from JavaScript Object Notation (JSON) manifests
rather than hand-transcribed.
Commands, policy manifests, compile-time constants, source identities,
hardware metadata, and evidence checksums are collected in
Appendix~\ref{app:reproduction}, together with the public code release.
Historical format paths and rejected timing
experiments remain in the other appendices so they do not define the primary
comparison.

\section{Results}

Throughput is reported in trillions or quadrillions of floating-point
operations per second (TFLOP/s or PFLOP/s).

\subsection{Operator performance and accuracy}

How much forward speed does Direct-P gain, and what accuracy does it trade for
that speed?  Table~\ref{tab:primary-results} places the same finite NV/MX route
on our GB200 and B300 systems and adds HAO's published GB300 NV/FP8 result
where the shape matches.  The complete local D128 grid underlies
Figures~\ref{fig:headline-pareto} and \ref{fig:cross-shape}; its generated
table remains in the release artifacts.  Unbounded shiftless NV/NV timings are
excluded even when a particular input happens to be finite.

\begin{table}[htbp]
\centering
\caption{Primary forward comparison across Blackwell systems.  Bold marks
independently confirmed B300 results above 3 PFLOP/s.  Published HAO columns
provide cross-run context.  ``ms / TF'' means milliseconds and TFLOP/s; TK
error is cosine / relative-$L_2$ against BF16.}
\label{tab:primary-results}
\begingroup
\small
\setlength{\tabcolsep}{4pt}
\begin{adjustbox}{max width=\textwidth}
\begin{tabular}{lccccc}
\toprule
Shape & TK GB200 ms / TF & TK B300 ms / TF & HAO GB300 NV/FP8 TF / cos.
& B300 $\Delta$t & TK B300 cosine / rel.-$L_2$ \\
\midrule
D128/H24/S4096 & 0.092160 / 2237 & 0.087008 / 2369 & 2046 / 0.9898 & -5.59\% & 0.9438 / 0.3363 \\
D128/H64/S4096 & 0.202752 / 2711 & 0.189536 / 2901 & -- & -6.52\% & 0.9440 / 0.3355 \\
D128/H64/S6144 & 0.452848 / 2731 & 0.418107 / 2958 & -- & -7.67\% & 0.9432 / 0.3381 \\
D128/H64/S8192 & 0.758336 / 2900 & \textbf{0.705684 / 3116} & -- & -6.94\% & 0.944063--0.944113 / 0.334931--0.335152 \\
D128/H64/S9472 & -- & \textbf{0.930724 / 3159} & -- & -- & 0.943960--0.944056 / 0.335198--0.335473 \\
D128/H24/S32768 & 4.400448 / 2998 & 4.480736 / 2945 & 2677 / 0.9899 & +1.82\% & 0.9429 / 0.3389 \\
\bottomrule
\end{tabular}
\end{adjustbox}
\endgroup
\end{table}

Across the nine GB200 D128 rows, \code{fast} is
\HaoGridFastGeoSpeedup$\times$ faster than HAO BF16 geometrically and peaks
at \HaoGridFastPeakTflops\ TFLOP/s. \code{Accurate} reaches
\HaoGridAccurateGeoSpeedup$\times$ and \HaoGridAccuratePeakTflops\ TFLOP/s.
At S32768/H24, fast takes 4.400448 ms (2998 TFLOP/s), compared with HAO's
published 2018 TFLOP/s on B200 and 2677 TFLOP/s on GB300 for NV/FP8.

HAO NV/FP8 remains more accurate: its mean cosine is
\HaoPublishedMeanCosine, versus \HaoGridFastMeanCosine\ for fast and
\HaoGridAccurateMeanCosine\ for accurate. The corresponding TK mean
relative-$L_2$ values are \HaoGridFastMeanRelLTwo\ and
\HaoGridAccurateMeanRelLTwo. HAO does not publish relative-$L_2$ for these
rows. Appendix~\ref{app:accuracy-matched-control} measures the cost of matching
its cosine with our exact NV/FP8 route; the local HAO NV/NV and full format
matrix remain in Appendices~\ref{app:nvnv} and \ref{app:format-matrix}.

B300 improves D128 latency by 5.6--7.7\% on the S4096--S8192 rows.
The standard S8192/H64 case reaches \BThreeThreeKStandardTflops\ TFLOP/s,
and wave-aligned S\BThreeThreeKPeakSeq/H64 reaches
\BThreeThreeKPeakTflops\ TFLOP/s. The exception is S32768/H24: B300 reaches
2945 TFLOP/s versus 2998 on GB200. At HAO's published D128 shapes, our B300
NV/MX reaches 2369 versus 2046 TFLOP/s at S4096/H24 and 2945 versus 2677 at
S32768/H24. These are different implementations and harnesses, so they
establish scale rather than causal timing differences. The distinct D64 tile
regime is reported in Appendix~\ref{app:nvnv-d64}.

\subsection{Speed--accuracy behavior across shapes}

\begin{figure}[htbp]
\centering
\includegraphics[width=.83\textwidth]
  {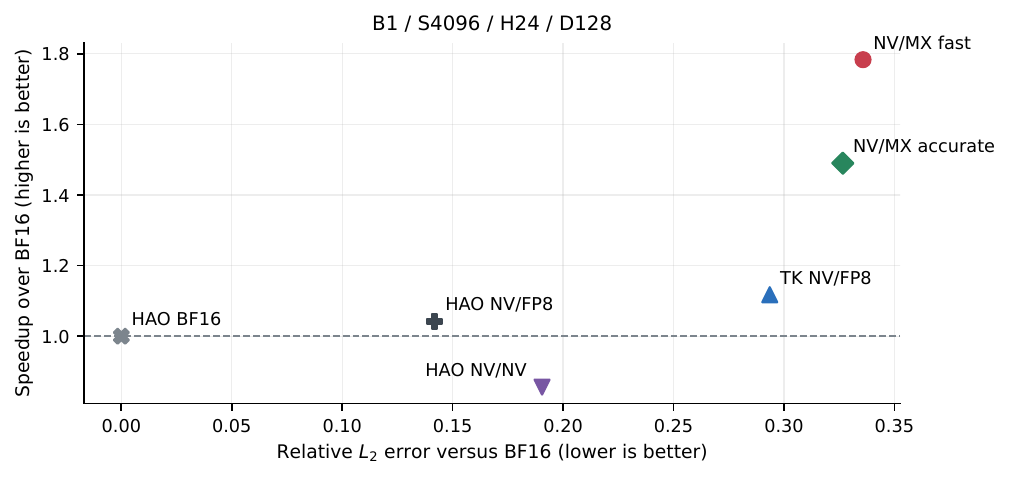}
\caption{Speed--error plane for the headline shape. Up and left are better.
The fastest full-FP4 points accept more operator error than HAO NV/NV or FP8
PV.}
\label{fig:headline-pareto}
\end{figure}

Does the trade-off persist across attention shapes?  Figure
\ref{fig:headline-pareto} shows the independent suite at $B=1$, $S=4096$,
$H=24$, and $D=128$, including optimized local FP8-PV controls. Its six-order
reference harness produces slightly different timings from
Table~\ref{tab:primary-results}, but all points share one BF16 value.

FP8 PV avoids the E2M1 payload and block-scale page required by full FP4 and
is more accurate. The local TK NV/FP8 control reaches only 1.118$\times$ BF16
in this suite, whereas NV/MX fast reaches 1.783$\times$; HAO's stronger
published NV/FP8 result is therefore included in the primary table rather
than inferred from this local control.

\begin{figure}[htbp]
\centering
\includegraphics[width=.96\textwidth]
  {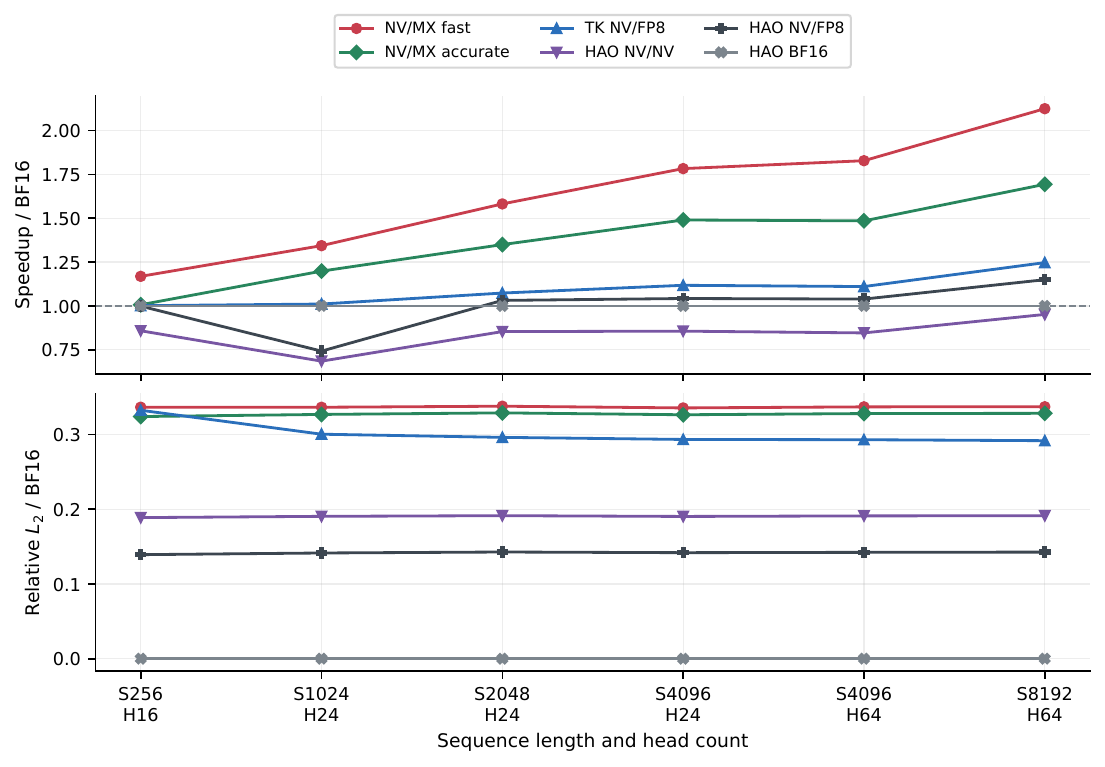}
\caption{BF16 speedup and relative-$L_2$ from matched records across startup,
transition, and saturated shapes.}
\label{fig:cross-shape}
\end{figure}

Figure~\ref{fig:cross-shape} adds H64 cases. At S4096/H64, fast takes
0.202752 ms versus 0.370688 ms for BF16 (1.828$\times$). At S8192/H64 it
takes 0.758336 ms versus 1.611488 ms (2.125$\times$). The two-query CTA is
important here: the earlier one-query topology created twice as many jobs and
could require an additional scheduling wave at high head count.

\Needspace{10\baselineskip}
\subsection{ViT and BERT fixed-input evaluations}

How does operator error propagate through complete transformer models?
Table~\ref{tab:downstream-tradeoff} joins physical attention speed, mean layer
error, and final task behavior. ViT uses 1,000 examples at S256 and 200 at
S1024/S4096. BERT MLM uses 800 S256 blocks and 200 S512 blocks; SST-2 uses all
872 validation examples.

\begin{table}[htbp]
\centering
\caption{Downstream fixed-input trade-off.  Task score is provider / BF16 accuracy;
final error is cosine / relative-$L_2$ against BF16.  Speedup belongs to the
physical attention shape, not the complete model.  HAO NV/NV is the
identical-input control because no NV/FP8 task evaluation is published.}
\label{tab:downstream-tradeoff}
\footnotesize
\begin{adjustbox}{max width=\textwidth}
\begin{tabular}{llrrrr}
\toprule
Task & Provider & Speedup & Task score / BF16 (\%)
& Final cos. / rel.-$L_2$ & $\Delta$MLM loss \\
\midrule
ViT S256 & TK NV/MX \code{fast} & 1.169$\times$ & 98.80 / 98.90 & 0.9987 / 0.0514 & -- \\
ViT S256 & TK NV/MX \code{accurate} & 1.007$\times$ & 98.70 / 98.90 & 0.9990 / 0.0452 & -- \\
ViT S256 & HAO NV/NV & 0.858$\times$ & 98.70 / 98.90 & 0.9997 / 0.0238 & -- \\
ViT S1024 & TK NV/MX \code{fast} & 1.344$\times$ & 95.00 / 96.50 & 0.9971 / 0.0768 & -- \\
ViT S1024 & TK NV/MX \code{accurate} & 1.198$\times$ & 97.00 / 96.50 & 0.9984 / 0.0564 & -- \\
ViT S1024 & HAO NV/NV & 0.685$\times$ & 97.00 / 96.50 & 0.9995 / 0.0324 & -- \\
ViT S4096 & TK NV/MX \code{fast} & 1.783$\times$ & 88.50 / 88.50 & 0.9887 / 0.1509 & -- \\
ViT S4096 & TK NV/MX \code{accurate} & 1.490$\times$ & 89.00 / 88.50 & 0.9989 / 0.0479 & -- \\
ViT S4096 & HAO NV/NV & 0.856$\times$ & 88.00 / 88.50 & 0.9992 / 0.0396 & -- \\
BERT SST-2 S256 & TK NV/MX \code{fast} & 1.169$\times$ & 92.32 / 92.43 & 0.9963 / 0.0870 & -- \\
BERT SST-2 S256 & TK NV/MX \code{accurate} & 1.007$\times$ & 92.09 / 92.43 & 0.9967 / 0.0815 & -- \\
BERT SST-2 S256 & HAO NV/NV & 0.858$\times$ & 92.20 / 92.43 & 0.9991 / 0.0431 & -- \\
BERT MLM S256 & TK NV/MX \code{fast} & 1.169$\times$ & 61.26 / 61.74 & 0.9968 / 0.0796 & +0.035 \\
BERT MLM S256 & TK NV/MX \code{accurate} & 1.007$\times$ & 61.36 / 61.74 & 0.9971 / 0.0758 & +0.028 \\
BERT MLM S256 & HAO NV/NV & 0.858$\times$ & 61.60 / 61.74 & 0.9983 / 0.0577 & +0.016 \\
BERT MLM S512 & TK NV/MX \code{fast} & 1.344$\times$ & 61.10 / 61.79 & 0.9969 / 0.0782 & +0.037 \\
BERT MLM S512 & TK NV/MX \code{accurate} & 1.198$\times$ & 61.36 / 61.79 & 0.9971 / 0.0761 & +0.040 \\
BERT MLM S512 & HAO NV/NV & 0.685$\times$ & 61.28 / 61.79 & 0.9982 / 0.0604 & +0.028 \\
\bottomrule
\end{tabular}
\end{adjustbox}
\end{table}

Both NV/MX policies remain finite. On ViT S4096, fast matches BF16 top-1
accuracy (88.5\%) with 95.5\% prediction agreement; accurate reaches 89.0\%
and 98.5\%, while native HAO NV/NV reaches 88.0\% and 98.0\%. Fast improves
from roughly 0.944 cosine in the Gaussian operator test to 0.9961 mean layer
cosine on this model input. Residual paths, normalization, later mixing, and
decision margins can attenuate operator error, so standalone relative-$L_2$
is informative but is not itself a task loss.

Across \DownstreamClassificationSamples\ classification examples, fast
changes \DownstreamFastPredictionChanges\ predictions, with
\DownstreamFastLowMarginChanges\ in the lowest quartile of BF16 top-two logit
margins. All changes from accurate and HAO NV/NV also lie in that quartile.
This supports a margin mechanism, but the suite is too small to establish
universal inference or training safety.

\subsection{Wan video diffusion}

Does Direct-P remain finite when attention error is composed across many
diffusion steps?  Wan is a larger, native-shape test: all video self-attention
calls run at S7680/D128, while text cross-attention and the rest of the model stay BF16.
Table~\ref{tab:wan-downstream} compares every route with paired HAO CuTe-DSL
BF16 outputs and warmed kernel timing.

\begin{table}[htbp]
\centering
\caption{Wan2.1 quality and warmed GB200 kernel speed at S7680/D128. Speedup
is relative to HAO CuTe-DSL BF16. Quality is cosine / relative-$L_2$ of the
final latent against the paired BF16 run.}
\label{tab:wan-downstream}
\scriptsize
\begin{adjustbox}{max width=\textwidth}
\begin{tabular}{llrrccc}
\toprule
Model & Method & Time (ms) & Speedup & 1 step & 4 steps & 20 steps \\
\midrule
Wan2.1-1.3B & HAO CuTe BF16 & 0.2888 & 1.00$\times$ & 1.0000 / 0.0000 & 1.0000 / 0.0000 & 1.0000 / 0.0000 \\
Wan2.1-1.3B & TK NV/MX fast & 0.1653 & 1.75$\times$ & 0.9870 / 0.1803 & 0.9671 / 0.2611 & 0.9136 / 0.4163 \\
Wan2.1-1.3B & TK NV/MX accurate & 0.1961 & 1.47$\times$ & 0.9862 / 0.1721 & 0.9654 / 0.2637 & 0.9060 / 0.4323 \\
Wan2.1-1.3B & HAO NV/NV & 0.3466 & 0.83$\times$ & 0.9878 / 0.1751 & 0.9711 / 0.2405 & 0.9389 / 0.3463 \\
Wan2.1-1.3B & HAO NV/FP8 & 0.2888 & 1.00$\times$ & 0.9936 / 0.1184 & 0.9767 / 0.2158 & 0.9530 / 0.3066 \\
\addlinespace
Wan2.1-14B & HAO CuTe BF16 & 0.8882 & 1.00$\times$ & 1.0000 / 0.0000 & 1.0000 / 0.0000 & 1.0000 / 0.0000 \\
Wan2.1-14B & TK NV/MX fast & 0.4250 & 2.09$\times$ & 0.9938 / 0.1179 & 0.9338 / 0.3589 & 0.8496 / 0.5337 \\
Wan2.1-14B & TK NV/MX accurate & 0.5072 & 1.75$\times$ & 0.9879 / 0.1563 & 0.9243 / 0.3946 & 0.8447 / 0.5500 \\
Wan2.1-14B & HAO NV/NV & 0.9073 & 0.98$\times$ & 0.9908 / 0.1358 & 0.9327 / 0.3640 & 0.9036 / 0.4435 \\
Wan2.1-14B & HAO NV/FP8 & 0.7516 & 1.18$\times$ & 0.9893 / 0.1464 & 0.9229 / 0.3877 & 0.8398 / 0.5669 \\
\end{tabular}
\end{adjustbox}
\end{table}

Fast is 1.75$\times$ faster than CuTe BF16 on the 1.3-billion-parameter model
and 2.09$\times$ on the 14-billion-parameter model; accurate reaches
1.47$\times$ and 1.75$\times$. HAO NV/FP8 is at parity for H12 and
1.18$\times$ faster for H40. On the 14-billion-parameter model at four steps,
fast reaches 0.9338 / 0.3589, close to HAO NV/NV at 0.9327 / 0.3640. At twenty
steps it reaches 0.8496 / 0.5337, versus 0.9036 / 0.4435 for HAO NV/NV; a
held-out prompt reaches 0.8235 / 0.5813. The route is finite but accumulates
more drift.

The original layer-39 failure was not an anchor miss: the sampled and exact
row maxima were equal. The expression $(s/6)\sum_i c_i$ formed a subnormal
intermediate at E8M0 code 1 and flushed to zero. Algorithm~\ref{alg:sampled-guard}
reassociates it as $s(\sum_i c_i/6)$. Both 14-billion-parameter twenty-step
runs now complete all 1,600 attention calls without a scan or stable-softmax
fallback.

Guarded layers are 21--23\% slower individually, but they are only 3/30 layers
in the 1.3-billion-parameter model and 4/40 in the 14-billion-parameter model.
Layer-weighted times are therefore 0.165309 ms and 0.424995 ms, 2.2--2.3\%
above the unguarded bases. The historical layer-wise affine map is omitted
because its small gain reversed between prompts. These drop-in BF16-checkpoint
results do not reproduce Attn-QAT's trained quality; that requires its QAT
checkpoint or training recipe \cite{zhang2026attnqat}.

\subsection{Paired image reconstruction}

Is the residual attention error visible in reconstructed images?
Table~\ref{tab:mae-reconstruction} replaces all 12 ViT-MAE encoder attention
layers for the same 100 images and masks. The S256 speedup is a physical kernel
measurement, not end-to-end MAE speed.

\begin{table}[htbp]
\centering
\caption{Paired ViT-MAE reconstruction. PSNR $\Delta$ is FP4 minus BF16 with
a paired 95\% interval. Reconstruction and layer cells are cosine /
relative-$L_2$ against BF16.}
\label{tab:mae-reconstruction}
\scriptsize
\begin{adjustbox}{max width=\textwidth}
\begin{tabular}{lrrrrrr}
\toprule
Provider & S256 speedup & PSNR (dB) & PSNR $\Delta$ (dB)
& MSE $\Delta$ (\%) & Reconstruction & Mean layer \\
\midrule
BF16 & 1.000$\times$ & 23.917 & -- & +0.00 & 1.00000 / 0.0000 & 1.0000 / 0.0000 \\
HAO NV/FP8 & 0.999$\times$ & 23.910 & -0.007 $\pm$ 0.012 & +0.16 & 0.99995 / 0.0089 & 0.9938 / 0.1046 \\
HAO NV/NV & 0.858$\times$ & 23.906 & -0.010 $\pm$ 0.015 & +0.07 & 0.99992 / 0.0113 & 0.9868 / 0.1589 \\
TK NV/MX \code{accurate} & 1.007$\times$ & 23.893 & -0.024 $\pm$ 0.021 & +0.25 & 0.99978 / 0.0184 & 0.9676 / 0.2616 \\
TK NV/MX \code{fast} & 1.169$\times$ & 23.896 & -0.020 $\pm$ 0.026 & +0.29 & 0.99973 / 0.0203 & 0.9600 / 0.2928 \\
\bottomrule
\end{tabular}
\end{adjustbox}
\end{table}

\begin{figure}[htbp]
\centering
\includegraphics[width=.98\textwidth]
  {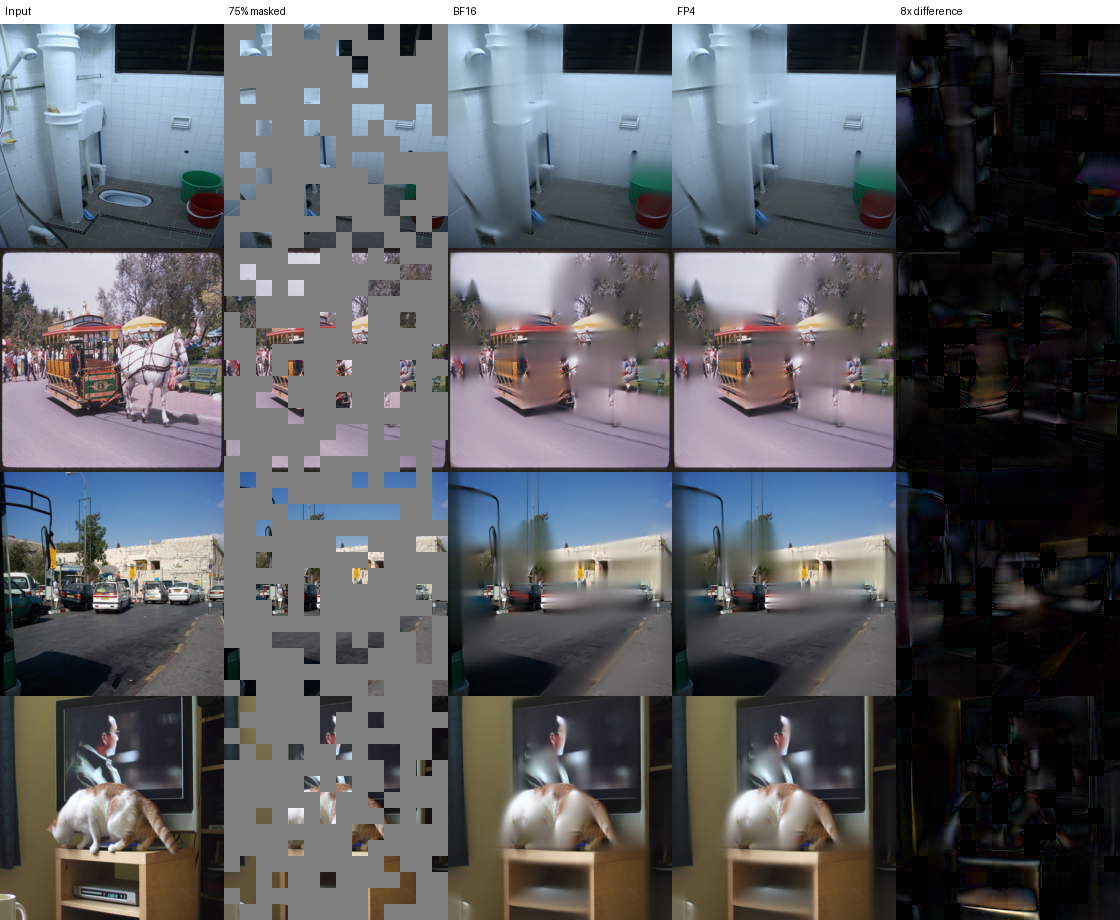}
\caption{Four paired ViT-MAE examples. The right column shows
$8\lvert I_{\mathrm{FP4}}-I_{\mathrm{BF16}}\rvert$; only encoder attention
differs between the BF16 and fast runs.}
\label{fig:mae-reconstruction}
\end{figure}

HAO NV/FP8 is closest to BF16 at $-0.007\pm0.012$ dB. HAO NV/NV, accurate,
and fast reach $-0.010\pm0.015$, $-0.024\pm0.021$, and
$-0.020\pm0.026$ dB. Fast has the largest reconstruction displacement, but
its output remains 0.99973 cosine and 2.03\% relative-$L_2$ from BF16 after
all twelve layers. Figure~\ref{fig:mae-reconstruction} makes the residual
visible only after an $8\times$ amplification.

The aggressive shiftless TK NV/NV control fails on the first sample of every
task, whereas HAO's stabilized NV/NV remains finite. Appendix~\ref{app:nvnv-failure}
shows that the distinction is stabilization, not the use of NVFP4 P itself.

\FloatBarrier
\section{Causal Training from Saved Quantized Operands}
\label{sec:causal-training}

Training adds three constraints to the forward problem.  Causal attention
masks future tokens, grouped-query attention (GQA) lets several query heads
share key/value heads, and backward must recover gradients without
materializing the full $S\times S$ score or probability matrices.  We first
present the retained method, then measure isolated backward,
projection-inclusive attention, a complete model update, and distributed
training.  Rejected designs remain in
Appendix~\ref{app:causal-design-history}.

\subsection{Pass the forward quantization into backward}

The forward pass saves the NVFP4 query/key payload bytes, their block scales,
the per-head Q--K global factors, and each row's log-sum-exp (LSE)
normalizer.  Backward recomputes $QK^\mathsf{T}$ from those bytes and scales,
then uses LSE to reconstruct the probabilities.  Separately, the projection
epilogue publishes row- or column-oriented FP8 views of Q, K, and V so each
gradient matrix product receives its required physical layout without a
standalone transpose or quantization launch.  Thus ``exact'' below means
exact with respect to the quantized forward payload, not to unquantized BF16
inputs.  Row-oriented and column-oriented describe how the same represented
values and scales are arranged for different matrix instructions; they do not
introduce a second quantization rule.

The implementation dispatches the Llama-style
8-billion-parameter attention shape at local batch sizes
$B\in\{1,2,4\}$, sequence length $S=4096$, 32 query heads, eight key/value
heads, and head dimension $D=128$.
The combined
query/key/value (QKV) projection applies rotary positional embedding (RoPE) and
publishes the layouts needed by forward and backward while its output fragments
are still live.  This avoids launching separate quantization and transpose
kernels around every attention layer.

This is not pure-FP4 end-to-end training.  Learned projections and attention
are separate precision boundaries.  We evaluate E4M3 projections as a
numerical control and NVFP4 projections as the higher-throughput arm.  Inside
attention, the controlled format comparison selects FP8 rather than MXFP4 for
P and V.  The supporting ablations are collected in
Appendix~\ref{app:causal-design-history}.

We prefix a tensor with d to denote its gradient; dO is the gradient arriving
from the output projection.

\begin{table}[htbp]
\centering
\caption{Precision contract for causal training.  Learned projections and
attention operands are separate boundaries.}
\label{tab:causal-precision-contract}
\small
\begin{tabularx}{\textwidth}{p{.25\textwidth}p{.27\textwidth}X}
\toprule
Part of the model & Representation & Reason \\
\midrule
Learned QKV and output projections & E4M3 control or NVFP4 throughput arm;
FP32 accumulation & Separates projection error from the attention-format
comparison. \\
Forward score product & NVFP4 Q and K & Reuses two-dimensional scales over
16-value blocks along the inner dimension (row-by-K16). \\
Forward value product & FP8 P and V & Retained training route; the MXFP4
alternative is faster in isolation but diverges in the observed distributed
experiments. \\
Probability reconstruction in backward & Saved NVFP4 Q/K, block and global
scales, and LSE &
Reconstructs the same represented probability used by forward without storing
an $S\times S$ matrix. \\
Backward gradient products & E4M3 Q/K/V/P/dS; E5M2 dO & E4M3 supplies
precision; E5M2 supplies the range needed for the small output gradient dO. \\
Gradient outputs & FP32 accumulation, BF16 dQ/dK/dV & Keeps accumulation and
optimizer-facing gradients stable. \\
\bottomrule
\end{tabularx}
\end{table}
\FloatBarrier

\subsection{How the backward pass works}

We write dQ, dK, dV, dP, and dS for gradients of the corresponding attention
tensors.  The backward pass executes five tiled matrix products, beginning
with $QK^\mathsf{T}$ reconstructed from the saved quantized operands:
\begin{equation}
  dP=dO\,V^\mathsf{T},\qquad
  dV=P^\mathsf{T}dO,\qquad
  dQ=dS\,K,\qquad
  dK=dS^\mathsf{T}Q.
  \label{eq:causal-bwd-products}
\end{equation}
The softmax gradient is
\begin{equation}
  r_i=\langle O_i,dO_i\rangle,\qquad
  dS=P\odot(dP-r\mathbf 1^\mathsf{T}),
  \label{eq:causal-bwd-ds}
\end{equation}
so probability reconstruction and dP must both complete before dQ and dK can
begin.  Causal masking removes illegal tiles but also makes work ownership
uneven.

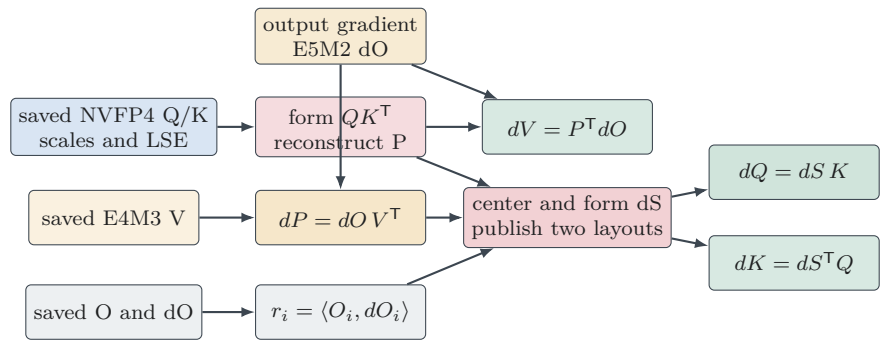
\begin{figure}[H]
\centering
\begin{tikzpicture}[
  font=\scriptsize,
  block/.style={draw=inkgray,rounded corners=2pt,align=center,
    minimum width=2.25cm,minimum height=.72cm,inner sep=3pt},
  edge/.style={-{Latex[length=1.8mm]},thick,draw=inkgray}]
  \node[block,fill=scaleyellow!20] (do) at (3.0,3.0) {output gradient\\E5M2 dO};
  \node[block,fill=scoreblue!18] (scorein) at (0,1.8)
    {saved NVFP4 Q/K\\scales and LSE};
  \node[block,fill=probred!18] (prob) at (3.0,1.8)
    {form $QK^\mathsf{T}$\\reconstruct P};
  \node[block,fill=outputgreen!18] (dv) at (6.0,1.8)
    {$dV=P^\mathsf{T}dO$};

  \node[block,fill=scaleyellow!14] (vin) at (0,.6) {saved E4M3 V};
  \node[block,fill=scaleyellow!24] (dp) at (3.0,.6)
    {$dP=dO\,V^\mathsf{T}$};
  \node[block,fill=probred!24] (ds) at (6.0,.6)
    {center and form dS\\publish two layouts};
  \node[block,fill=outputgreen!22] (dq) at (9.0,1.2)
    {$dQ=dS\,K$};
  \node[block,fill=outputgreen!18] (dk) at (9.0,0)
    {$dK=dS^\mathsf{T}Q$};

  \node[block,fill=lightgray] (oin) at (0,-.65) {saved O and dO};
  \node[block,fill=lightgray] (row) at (3.0,-.65)
    {$r_i=\langle O_i,dO_i\rangle$};

  \draw[edge] (scorein) -- (prob);
  \draw[edge] (prob) -- (dv);
  \draw[edge] (do) -- (dv);
  \draw[edge] (do) -- (dp);
  \draw[edge] (vin) -- (dp);
  \draw[edge] (prob) -- (ds);
  \draw[edge] (dp) -- (ds);
  \draw[edge] (oin) -- (row);
  \draw[edge] (row) -- (ds);
  \draw[edge] (ds) -- (dq);
  \draw[edge] (ds) -- (dk);
\end{tikzpicture}
\caption{Causal backward from saved quantized operands.  The serial dependency
runs from probability reconstruction and dP through dS to dQ/dK.  The dV
branch also consumes P and dO, while the saved output and dO provide the
row-centering statistic.  The
equations inside the output boxes name Q/K inputs that are not drawn as extra
crossing arrows.}
\label{fig:causal-backward-dag}
\end{figure}

The implementation accelerates this graph in four ways:
\begin{enumerate}
  \item reconstruct scores from the compact forward payload rather than
  creating a separate BF16 score path;
  \item reuse the rounded P representation for dV and dS instead of decoding
  it twice;
  \item publish both physical dS layouts from the producer fragment and
  release aliased tensor memory as soon as the next consumer owns it; and
  \item publish dO directly as E5M2.  E4M3 rounded roughly 97\% of the observed
  dO values to zero in the failing checkpoint diagnostic, whereas E5M2
  reduced the zero fraction to roughly 14\% with less than 1\% publisher
  overhead.
\end{enumerate}

The D128 backward remains schedule-limited and allows only one CTA per
streaming multiprocessor.  Detailed occupancy and activity counters exist only
for a predecessor schedule, so they are kept with that schedule's timings in
Appendix~\ref{app:causal-design-history} rather than attributed to the retained
binary.

\subsection{Isolated backward}

Does passing the saved quantization into backward make the kernel itself
faster?  We first time only causal attention backward at the production head
dimension,
$D=128$, on one GB200.  Both paths are prepared and bound before timing.  The
BF16 control decodes the same represented E4M3 Q, K, V, and E5M2 dO
operands, whereas the quantized path uses the NVFP4 Q/K payload saved by
forward.  Projection, allocation, loss, and optimizer work
are outside this boundary.

\begin{table}[htbp]
\centering
\caption{Isolated D128 causal backward at B1/S4096.  Latency is the median of
warmed runs; speedup is BF16 latency divided by the latency in each row.}
\label{tab:causal-backward-isolated}
\small
\begin{tabular}{lrr}
\toprule
Route & Latency (ms) & Speedup \\
\midrule
BF16 FA4 backward & 0.501 & 1.000$\times$ \\
Saved-Q/K reconstruction core & 0.356 & 1.405$\times$ \\
Core + E5M2 dO/statistics publisher & 0.508 & 0.986$\times$ \\
\bottomrule
\end{tabular}
\end{table}
\FloatBarrier

The reconstruction core saves 29\% of the BF16 backward time, but the
training-safe E5M2 dO and row-statistic publisher consumes almost exactly that
saving.  This is why a fast backward kernel alone does not predict an
end-to-end gain.  The core passes the exact-zero-dO test, returns finite
nonzero gradients for a nonzero dO, and passes its represented-operand
reference gate.  In the
paired projection-inclusive check, output and input-gradient cosine against
BF16 are 0.998 and 0.956; Q, K, V, and output-projection weight-gradient
cosines range from 0.983 to 0.999.  These are local operator checks, not a
convergence result.

\begin{figure}[H]
\centering
\reportplot[0.20\textheight]{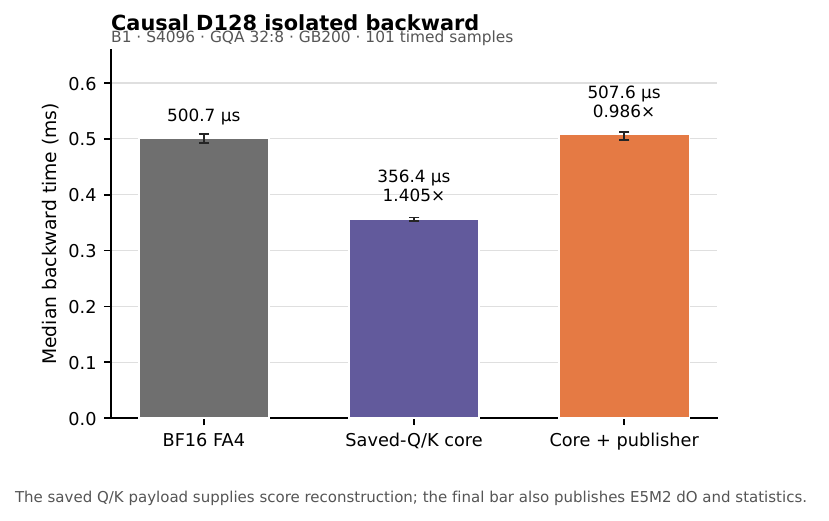}
\caption{Isolated D128 causal-backward latency for BF16 and the path that
reconstructs P from saved quantized Q/K at B1/S4096.}
\label{fig:causal-isolated-shape-speedups}
\end{figure}

\Needspace{10\baselineskip}
\subsection{Combined forward and backward}

Does that saving survive its producers and consumers?  The second boundary is
the complete attention sublayer: packed QKV projection,
rotary positional embedding, scale and layout publication, causal attention,
output projection, and their gradients.  It includes the producer and handoff
costs hidden by the isolated kernel test.  It excludes the surrounding root
mean square normalization (RMSNorm), residual connection, language-model
loss, and optimizer.  The BF16 and quantized paths use identical
projection weights and the same B1/S4096/D128 model shape.

\begin{table}[htbp]
\centering
\caption{Projection-inclusive attention on one GB200.  Backward-only runs the
same prepared forward outside the timing interval; the second row times both
forward and backward.  Values are medians of warmed runs.}
\label{tab:causal-combined-forward-backward}
\small
\begin{tabular}{lrrr}
\toprule
Boundary & BF16 (ms) & Quantized route (ms) & Speedup \\
\midrule
Backward only & 1.572 & 1.397 & 1.125$\times$ \\
Forward + backward & 2.656 & 2.133 & 1.245$\times$ \\
\bottomrule
\end{tabular}
\end{table}
\FloatBarrier

\begin{figure}[H]
\centering
\reportplot[0.17\textheight]{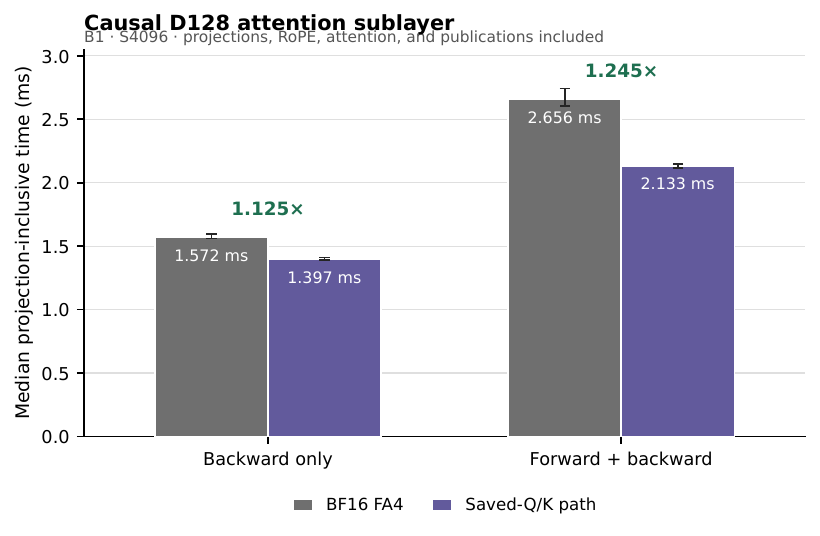}
\caption{Matched projection-inclusive attention forward and backward:
BF16 FA4 versus the saved-quantization path at B1/S4096/D128.}
\label{fig:causal-combined-forward-backward}
\end{figure}

\subsection{Eight-billion-parameter end-to-end result}

Does the attention gain remain visible in a complete model update?  We measure
a full update of a Llama-3.1-style 8.03-billion-parameter
model with 32 layers, 32 query heads, eight key/value heads, and head dimension
128.  We sweep local batch 1, 2, and 4 at sequence length 4096 on one GB200.
Every arm uses the same fused AdamW optimizer and standard dense cross entropy
compiled with \texttt{torch.compile}; cut cross entropy (CCE) is disabled.
Each process is bound to the CPU and memory node local to its GPU.  We report
the median after 10 warmups and 21 measured updates over fixed synthetic
tokens.

The quantized arms use NVFP4 QKV and output projections, row-by-K16 NVFP4
Q/K inside attention, and either FP8 or MXFP4 P/V.  Their backward binary is
identical at a given batch; their forward P/V implementations are
format-specific.  Each quantized measurement is paired with a freshly
measured packed-QKV BF16 FA4 control.

\begin{table}[htbp]
\centering
\caption{Complete 8B updates at S4096 on one GB200.  Each P/V route has its own
adjacent BF16 timing bracket.  Times are medians in milliseconds.}
\label{tab:llama8b-e2e}
\small
\begin{adjustbox}{max width=\textwidth}
\begin{tabular}{rrrrrrr}
\toprule
& \multicolumn{3}{c}{FP8 P/V bracket} & \multicolumn{3}{c}{MXFP4 P/V bracket} \\
\cmidrule(lr){2-4}\cmidrule(lr){5-7}
Local batch & BF16 & Low precision & Speedup & BF16 & Low precision & Speedup \\
\midrule
B1 & 260.313 & 239.985 & 1.085$\times$ & 261.133 & 239.250 & 1.091$\times$ \\
B2 & 464.245 & 415.532 & 1.117$\times$ & 463.814 & 415.408 & 1.117$\times$ \\
B4 & 854.516 & 751.722 & 1.137$\times$ & 857.226 & 751.597 & 1.141$\times$ \\
\bottomrule
\end{tabular}
\end{adjustbox}
\end{table}
\FloatBarrier

The benefit grows as the GPU is better filled: approximately 1.09$\times$ at
B1, 1.12$\times$ at B2, and 1.14$\times$ at B4.  In the FP8 bracket, B4
throughput rises from 19,173 to 21,795 tokens/s per GPU and measured model FLOP
utilization from 41.12\% to 46.74\%.  The MXFP4 arm reaches 21,799 tokens/s and
46.75\% utilization.  FP8 and MXFP4 are therefore tied end to end: their
quantized step times differ by at most 0.31\%, no larger than variation
between their separately measured BF16 anchors.  MXFP4 is modestly faster
in the timed forward portion, but that sub-millisecond difference is not a
material complete-update win.

\begin{figure}[tbp]
\centering
\reportplot[0.19\textheight]{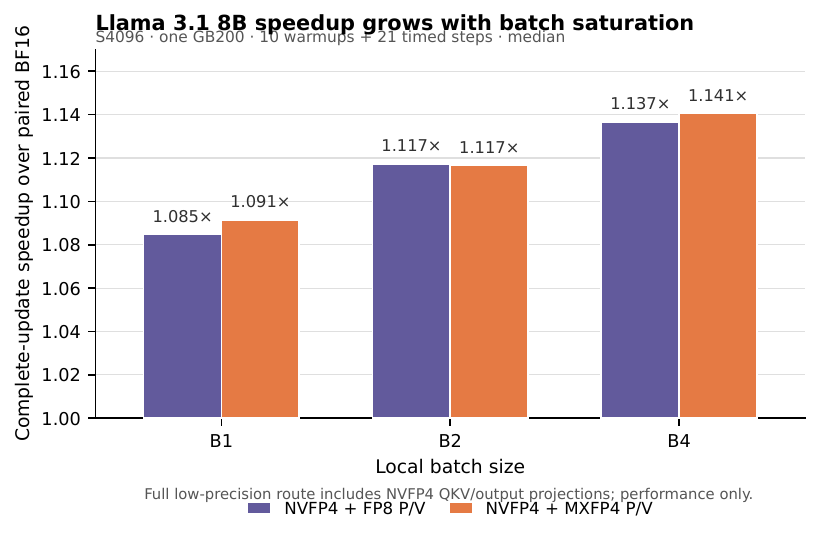}
\caption{Complete-update speedup for the two quantized routes as local
batch increases.  These full-route values include NVFP4 QKV/output projections
and therefore are not attention-only or P/V-only speedups.}
\label{fig:llama8b-e2e-breakdown}
\end{figure}

This matrix is deliberately \emph{performance-only}.  Initial-logit cosine
against BF16 is 0.416--0.426 for FP8 P/V and 0.373--0.374 for MXFP4 P/V;
relative-$L_2$ is approximately 1.07 and 1.12.  These short fixed-token updates
do not establish training quality.  The longer FP8 and MXFP4 trajectories in
Appendix~\ref{app:causal-design-history} are separate experiments; a fast
finite timing bracket must not be read as a convergence result.

\subsection{Training stability selects FP8 P/V}

Which probability format remains stable in longer training?  A historical
four-arm diagnostic crossed E4M3 and NVFP4 learned projections with FP8 and
MXFP4 P/V while holding the backward path fixed.  Both FP8-P/V arms remain
non-divergent and descend through their common observed horizon of 55.5
billion tokens.  Both MXFP4-P/V arms separate from their FP8 controls near
0.1 billion tokens and later develop rising loss and very large pre-clipping
gradient norms.  A subsequent matched B4 launch shows the same failure for
NVFP4 projections with MXFP4 P/V: it tracks FP8 through update 300, then its
loss rises from 7.01 at update 325 to 16.25 at update 350.  These experiments
select FP8 P/V for the retained training route.  The complete historical
curves and failure analysis are in
Appendix~\ref{app:causal-design-history}.

\subsection{Matched distributed training}

How do the retained route and BF16 compare at the same training coordinates?
The matched study uses 64 GPUs, local batch four, four gradient
accumulation steps, and effective global batch 1024.  Its BF16 and
NVFP4-projection/FP8-PV arms share the model, optimizer, tokenizer, sample
order, and token schedule.  Both trajectories support exact checkpoint
resume and complete the 100,000,595,968-token schedule.  The terminal update
saves the final checkpoint but is not a scheduled metric report.  We therefore
compare training at update 23,825 (99.93 billion tokens) and held-out
validation at update 23,840 (99.99 billion tokens), the last common scheduled
coordinates before the terminal update.

\begin{figure}[tbp]
\centering
\includegraphics[width=0.82\textwidth]{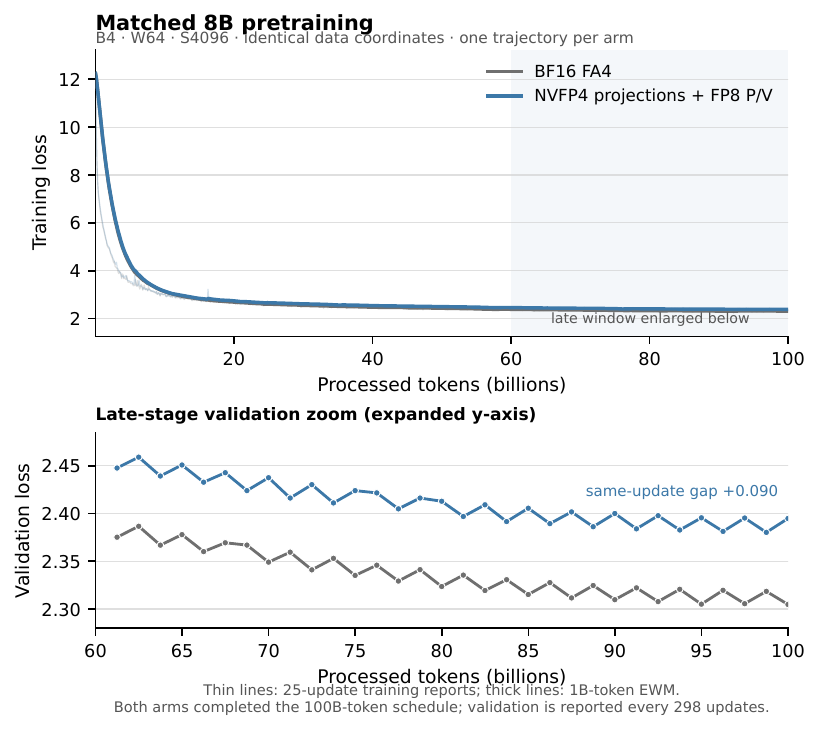}
\caption{Token-aligned training and held-out validation for the matched B4
experiment over the completed 100-billion-token schedule.  The panel compares
the BF16 control with the retained FP8-P/V route; the lower panel magnifies
same-update validation from 60 billion tokens onward with an expanded vertical
axis.  Each curve is one trajectory, so the plot provides no uncertainty
estimate.}
\label{fig:llama8b-b4-matched-training}
\end{figure}
\FloatBarrier

At the last scheduled training report, BF16 and the FP8 route report losses of
2.3095 and 2.3613 and pre-clipping gradient norms of 0.0549 and 0.0532,
respectively.  At the final same-update held-out validation report, the losses
are 2.3048 and 2.3948, a gap of 0.0900.
The FP8 trajectory is therefore stable and descending, but it is not
numerically identical to BF16.  Because the complete route changes both
the learned projections and attention, this comparison does not attribute the
gap to attention alone.  Across all 874 common logged coordinates from the end
of warmup through update 23,825, median throughput is 21,853
tokens/s/GPU for BF16 and 24,303 tokens/s/GPU for the FP8 route; the ratio
of medians is 1.112$\times$.  The paired throughput ratio has a 10th--90th
percentile range of 1.080--1.114$\times$.  This aggregation excludes no
observations, including input stalls and checkpoint windows.  This is one
trajectory per route and therefore does not estimate run-to-run variability.

\begin{figure}[H]
\centering
\reportplot[0.19\textheight]{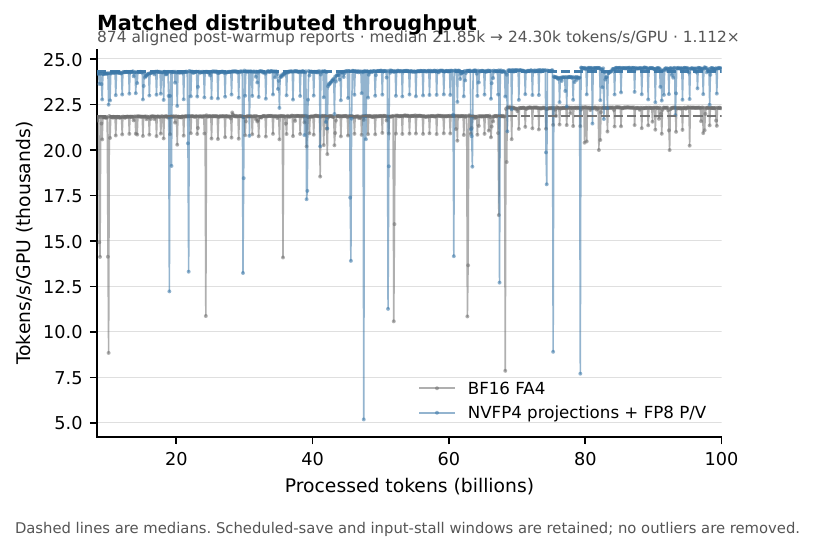}
\caption{Distributed throughput for the matched B4 experiment over
all 874 common post-warmup observations.  Input stalls and checkpoint windows
remain in the distribution.}
\label{fig:llama8b-b4-matched-throughput}
\end{figure}
\FloatBarrier

The three local boundaries establish isolated-backward,
projection-inclusive, and full-update speed at D128.  The distributed
evidence establishes the FP8-versus-MXFP4 format choice and includes a
same-recipe BF16 control with same-update validation.  It supports comparisons
over the completed 100-billion-token schedule, but not
statistical-equivalence claims or run-to-run uncertainty estimates.

\FloatBarrier
\section{Discussion and Conclusion}
\label{sec:current-headroom}

The forward, backward, and training results point to three hardware
takeaways.  We state each takeaway first and then give the measurement that
supports it; the rejected kernel variants remain in the appendices.

\begin{table}[htbp]
\centering
\caption{Hardware takeaways and the evidence that supports them.}
\label{tab:hardware-takeaways}
\small
\begin{tabularx}{\textwidth}{p{.28\textwidth}X}
\toprule
Takeaway & Supporting evidence \\
\midrule
Tensor-memory ownership limits overlap & Two FP32 score banks and two FP32
output accumulators use all $4\times128=512$ tensor-memory columns; reducing
shared-memory use did not increase CTA residency or reduce latency. \\
Readiness stalls leave matrix throughput unused & A matched diagnostic kept
the same 98,304 tensor instructions but reached only 18.8\% tensor-pipe
activity with the real probability path. \\
Probability arithmetic is no longer the main gap & A fixed-P diagnostic that
removes nearly all probability construction improved latency by only 5.23\%. \\
\bottomrule
\end{tabularx}
\end{table}

\subsection{Takeaway 1: TMEM capacity and ownership bound the overlap window}

The two-query pipeline in Figure~\ref{fig:quarter-pipeline} already overlaps
QK, softmax, and PV across query stages.  The limitation is more specific:
under the current D128 layout, the next QK cannot fully overlap the still-owned
probability and tail-PV phase.  Figure~\ref{fig:tmem-layout} shows why.  Two
128-column score banks and two 128-column FP32 output accumulators occupy all
512 tensor-memory (TMEM) columns.  A retired part of a score bank is reused for
the quantized probability and scale pages.  Until PV consumes that overlay,
the next QK targeting that score bank and query stage has no legal score
destination; the other query stage still provides partial overlap.

This is a storage-ownership dependency, not simply a shortage of barriers or
shared memory.  In a matched control, reducing dynamic shared memory from
209,920 to 163,840 bytes did not improve latency because the 512-column TMEM
allocation still allowed only one CTA per streaming multiprocessor.  More
shared memory, or more raw on-chip bytes that cannot be allocated as another
score bank, would not change this schedule.

\begin{figure}[htbp]
\centering
\begin{tikzpicture}[font=\small,
  bank/.style={draw=inkgray,minimum width=2.35cm,minimum height=.72cm,
    align=center},
  extra/.style={draw=scoreblue,dashed,minimum width=2.35cm,
    minimum height=.72cm,align=center}]
  \node[anchor=east,font=\bfseries] at (-.25,1.05) {Current};
  \node[bank,fill=scoreblue!22] (s0) at (1.0,1.05) {score 0 / P 0};
  \node[bank,fill=scoreblue!34,right=0pt of s0] (s1) {score 1 / P 1};
  \node[bank,fill=outputgreen!22,right=0pt of s1] (o0) {output 0};
  \node[bank,fill=outputgreen!34,right=0pt of o0] (o1) {output 1};
  \node[below=3pt of s1.south east,anchor=north,text=inkgray]
    {all 512 columns owned};

  \node[anchor=east,font=\bfseries] at (-.25,-.55) {Candidate contract};
  \node[bank,fill=scoreblue!22] (t0) at (1.0,-.55) {score 0 / P 0};
  \node[bank,fill=scoreblue!34,right=0pt of t0] (t1) {score 1 / P 1};
  \node[bank,fill=outputgreen!22,right=0pt of t1] (u0) {output 0};
  \node[bank,fill=outputgreen!34,right=0pt of u0] (u1) {output 1};
  \node[extra,fill=scoreblue!8,right=4pt of u1] (next) {next score};
  \node[below=4pt of t1.south,font=\scriptsize,text=scoreblue]
    {PV still consuming};
  \node[above=4pt of next.north,font=\scriptsize,text=scoreblue] (qkwrite)
    {next QK writes};
  \draw[-{Latex[length=2mm]},thick,scoreblue]
    (qkwrite.south) -- (next.north);
\end{tikzpicture}
\caption{Conceptual hardware implication.  The useful addition is another
allocatable score destination, together with legal issue and ownership
semantics, so the next QK can proceed while PV still consumes the current
overlay.  The lower row is not a measured kernel or a claim that raw capacity
alone is sufficient.}
\label{fig:extra-score-bank}
\end{figure}
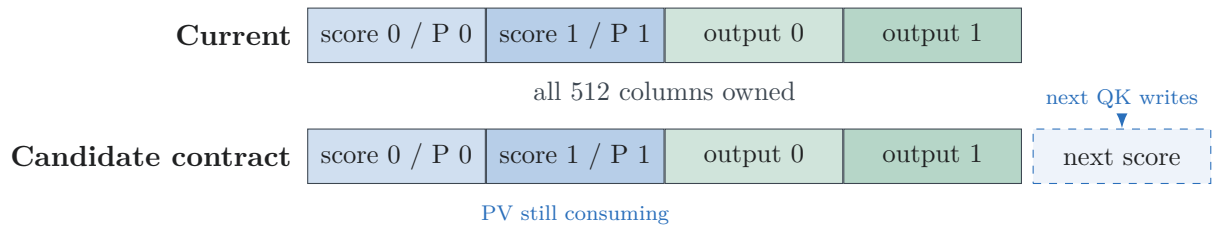

\subsection{Takeaway 2: low tensor activity is a readiness symptom}

The scaled-FP4 PV instruction consumes a K64 inner dimension, so two adjacent
32-column probability fragments must be ready before useful PV work can
start.  Publishing the first fragment alone does not advance the matrix
issuer.  The score-to-probability path therefore creates paired rendezvous:
Q0 and Q1 enable the first PV half, while Q2 and Q3 enable the tail.  Attempting
to issue the next QK between the first and tail PV operations violates the
current sequence of two K64 tensor-core accumulations.

Table~\ref{tab:historical-readiness-profile} shows a matched historical
profiling checkpoint.  The real and fixed-probability variants execute the
same number of tensor instructions.  Removing probability work shortens wall
time and raises the fraction of cycles attributed to the tensor pipe, but it
does not create more matrix work.  Source sampling attributes 1,715 of 2,669
not-issued samples to dependency stalls categorized by NVIDIA's profiler as
long-scoreboard waits, chiefly final statistics, score readiness, and output
publication.  Tensor-core underutilization is therefore a symptom of operands
not being ready, not a lack of nominal FP4 throughput.

\begin{table}[htbp]
\centering
\caption{Matched historical profile with identical tensor work.  This
diagnostic predates the final Direct-P binary and is used only to identify the
stall mechanism.}
\label{tab:historical-readiness-profile}
\small
\begin{tabular}{lrrr}
\toprule
Probability path & Dynamic instructions & Tensor instructions & Tensor active \\
\midrule
Real probability construction & 54.6 million & 98,304 & 18.8\% \\
Fixed probability & 11.9 million & 98,304 & 26.6\% \\
\bottomrule
\end{tabular}
\end{table}

The final Direct-P path has already removed most exposed probability cost.
At B1/S4096/H24/D128, the retained kernel measures 0.092448 ms and repeats at
0.092512 ms.  Intentionally incorrect ceilings bound what remains:

\begin{table}[htbp]
\centering
\caption{Final-kernel ceilings relative to the 0.092448-ms valid record.  These
rows deliberately remove required work and do not compute valid attention.}
\label{tab:speed-of-light}
\small
\begin{tabular}{lrr}
\toprule
Diagnostic & Time (ms) & Gap from 0.092448 ms \\
\midrule
Simplified score packing & 0.091168 & 1.280\us\ (1.38\%) \\
Keep row maximum, pack raw scores & 0.090112 & 2.336\us\ (2.53\%) \\
Use a fixed probability tile & 0.087616 & 4.832\us\ (5.23\%) \\
\bottomrule
\end{tabular}
\end{table}

Even the fixed-P diagnostic that removes nearly all probability construction
saves only 5.23\%.  The remaining gap to the four-times-BF16 matrix ratio
includes score loads, K64 publication granularity, scale delivery, online
correction, output accumulation, and the epilogue.  This is also why fewer
arithmetic instructions do not always help: independent arithmetic can occupy
issue slots while the kernel is waiting for a dependency.

\subsection{Takeaway 3: faster arithmetic does not scale the whole kernel}

\begin{table}[htbp]
\centering
\caption{Hardware properties relevant to the measured kernels.  TMEM capacity
does not increase from GB200 to the tested B300.}
\label{tab:blackwell-generations}
\small
\begin{tabular}{lcc}
\toprule
Property & GB200 / SM100 & B300 / SM103 \\
\midrule
Visible SMs in this study & 152 & 148 \\
Maximum reported clock & 2062 MHz & 2032 MHz \\
TMEM per SM & 256 KB & 256 KB \\
Key exponential rate & 16 ops/clock/SM & 32 ops/clock/SM \\
Dense NVFP4 GPU class & 1.0$\times$ & 1.5$\times$ \\
Fused TMEM load and reduction & no & yes \\
\bottomrule
\end{tabular}
\end{table}

Blackwell Ultra increases dense NVFP4 throughput, doubles important
special-function rates, and adds a fused TMEM load/reduction instruction
\cite{nvidia2025blackwellultra,nvidia2026ptx,zadouri2026flashattention4}.
Those improvements do not enlarge the score/output allocation, so the whole
attention loop does not become 1.5$\times$ faster.  The measurements are
consistent with faster QK and PV making score reduction, scale publication,
and correction more visible.

Launch shape matters as well.  A persistent grid wider than the 148 visible
B300 SMs creates a mostly empty second wave; recompiling for the device removes
that tail.  With the corrected geometry, S8192/H64 and the wave-aligned
S9472/H64 cases sustain 3116 and 3159 TFLOP/s.  At S32768/H24, however, B300
reaches 2945 TFLOP/s versus 2998 on GB200.  Per-SM-clock normalization favors
B300, but that normalization is an inference rather than an observed
GPU-throughput win.

Noncausal D64 measurements and unsafe-format controls are reported in
Appendix~\ref{app:nvnv-d64}.  Their different tile and ownership regime should
not be inferred from the D128 hardware profile above.

\subsection{Implications for hardware and kernels}

The measured dependencies motivate these candidate changes:
\begin{description}
  \item[Another allocatable score bank.] Allow the next QK to write while PV
  still owns the current probability overlay.  This must preserve two-query
  K/V reuse and provide legal issue semantics; a naive ping-pong prototype
  doubled the query jobs and was 52.8\% slower.
  \item[K32 scaled-FP4 PV.] Consume each 32-column probability completion
  immediately instead of waiting for a K64 pair.
  \item[Scales outside TMEM.] Remove instruction-facing scale pages from the
  overlay by reading compact scales from shared memory or registers.  Existing
  scale-footprint probes show that this alone neither creates a full
  128-column score destination nor demonstrates a speedup.
  \item[Wider tiles with a compatible lifecycle.] B300's wider matrix
  instructions have high synthetic ceilings, but only if probability
  production, K/V reuse, and score storage can be scheduled together.
\end{description}

These are hypotheses motivated by measured dependencies, not measured wins.
The resulting design implication is that, once Direct-P shortens probability
construction, another polynomial approximation is likely less valuable than
a larger usable overlap window.

\subsection{Scope and limitations}

The noncausal model experiments are fixed-input inference evaluations rather
than finetuning or pretraining studies.  The causal measurements establish
operator, complete-sublayer, and full-update performance.  The matched
distributed comparison supports training and validation claims over the
completed 100-billion-token schedule in Section~\ref{sec:causal-training};
one trajectory per route does not estimate run-to-run variability.  The longer historical
controls establish the FP8-versus-MXFP4 P/V format choice, but their different
recipe prevents a paired quality comparison with BF16.  Learned projection
precision is also a separate boundary: E4M3 is the numerical control and
NVFP4 is the measured throughput arm.

The hardware conclusions are shape-specific.  Head dimension 64 and other
regimes require their own tile size, cooperative-thread-array ownership, and
tensor-memory overlap strategy; they should not inherit the D128 schedule by
analogy.

\subsection{Conclusion}

Direct-P makes full-FP4 forward attention useful by shortening the serial path
between the score and value matrix products.  It maps log-softmax scores
directly to MXFP4 codes and normalizes with the same represented probabilities
consumed by the value product.  The result exceeds twice the BF16 forward
throughput on favorable GB200 shapes.  The FP8 probability route remains more
accurate, while the fixed-input model evaluations identify cases in which the
additional Direct-P error remains small at the model output.

For causal training, the forward pass supplies backward with its quantized
Q/K payload, scales, and softmax normalizer.  Projection and gradient
epilogues also publish the row- and column-oriented FP8 views required by the
gradient products.  This path accelerates the complete projection-inclusive
attention sublayer by 1.25$\times$ and a single-GPU 8B update by up to
1.14$\times$.  The distributed controls select FP8 P/V: every tested MXFP4
P/V trajectory diverges despite competitive short timing runs.

The retained training path therefore still makes a numerical concession to
FP8 for P/V and backward.  Unsigned E5M3 (UE5M3) block scales retain E2M1
payloads but provide substantially more scale range, and have enabled stable
FP4 language model pretraining in separate work~\cite{hu2026ue5m3}.  Applying
that format to the P/V product and backward gradient products is a promising
route to a fully FP4 attention training path, but it is not evaluated here and
requires an efficient hardware implementation.

Faster matrix arithmetic alone is therefore insufficient.  At head dimension
128, tensor-memory ownership and operand readiness limit how much QK, softmax,
and PV can overlap.  Low-precision attention needs a larger allocatable
overlap window as much as it needs faster tensor cores.

\section*{Generative-AI Disclosure}
Generative-AI tools assisted with code development, experiment orchestration,
data analysis, figure generation, and manuscript drafting and editing.  The
author reviewed the generated material, verified the reported measurements
against the cited repository artifacts, and takes responsibility for the
paper.

\bibliographystyle{plainnat}
\bibliography{main}

\appendix
\section{Historical NV/NV Investigation}
\label{app:nvnv}

This appendix preserves the NV/NV work that led to the retained design. It
is intentionally separated from the principal result because those kernels
use a different P scale format, approximation policy, and sometimes a
different denominator. Their timings must not be mixed with the final NV/MX
headroom.

\subsection{Initial stabilized path}

The first numerically robust full-FP4 route followed the conventional order:
\begin{equation}
  z \rightarrow m \rightarrow e^{z-m} \rightarrow a_B
  \rightarrow s_{\mathrm{E4M3}} \rightarrow \mathrm{E2M1}.
\end{equation}
It computed an exact online row maximum, produced floating probabilities,
reduced an N32 block maximum, encoded an E4M3 scale, divided by that scale,
and packed E2M1. This path remained finite on downstream inputs but measured
about 0.1884\,ms at S4096/H24. The sequence was numerically conservative and
serial: scale selection could not start until floating probabilities existed.

For a global NV encode factor $G$, payload and scale can be expressed as
\begin{align}
  S_B(G) &= Q_{\mathrm{E4M3}}\!\left(G\frac{a_B}{6}\right), \\
  q_j(G) &= \Qe\!\left(\frac{G p_j}{S_B(G)}\right), \\
  \widehat p_j(G) &= \frac{S_B(G)}{G}q_j(G).
\end{align}
If $G$ multiplies both numerator and denominator consistently and no
rounding boundary changes, it cancels. It is not a free accuracy knob once
scale rounding, underflow, and saturation are considered. Sweeping $G$
without changing the shiftless estimator produced little benefit; adding
full stabilization to make the sweep meaningful restored the expensive
path.

\subsection{Code-directed polynomial development}

The key useful idea from NV/NV was to target E2M1 decisions directly. A
cubic fit to the base-two locations of E2M1 thresholds was
\begin{equation}
  \begin{aligned}
    p(x)={}&0.07839806x^3+0.28625049x^2 \\
           &+0.63145205x+0.99202336.
  \end{aligned}
\end{equation}
After folding the score and block scale into its coefficients, two values
could be evaluated with packed Horner operations:
\begin{align}
  r_1 &= \operatorname{FMA2}(x,a,b), \\
  r_2 &= \operatorname{FMA2}(r_1,x,c), \\
  y &= \operatorname{FMA2}(r_2,x,d).
\end{align}
This was faster than a full exponential and more accurate than early linear
fits. It also established that the fitting target should be
$\Qe(f(x))$, not $f(x)$ itself.

An intermediate SFU/ALU hybrid issued four native \code{EX2} pairs first and
filled their latency with twelve cubic or affine pairs. The native samples
also supplied an approximate denominator. This improved over pure-SFU and
pure-ALU variants on that schedule, but it coupled numerator and denominator
sampling to one distribution.

The later affine endpoint
\begin{equation}
  \ell(x)=\max(0,1.62330034x+0.92083546)
\end{equation}
reduced non-native work to one packed FMA per pair. That mechanism survives
in the retained NV/MX path, with policy-specific coefficients and an exact
represented denominator.

\subsection{Sampled denominator}

The NV/NV throughput path sampled eight of 32 scalar probability values in
each quarter. If $\mathcal J_q$ is the rotated sample set, it estimated
\begin{equation}
  \widetilde L_B = A_B\,4
  \sum_{j\in\mathcal J_q}2^{x_j}.
\end{equation}
The factor four compensates for sampling one quarter of the values. This
saved a complete denominator pass but could miss a concentrated maximum or
misestimate a nonuniform tail. Two- and three-word denominator reductions
were faster diagnostics but failed downstream: in a 20-image ViT gate they
gave respectively 0\% and 75\% top-1 agreement with BF16. The retained
NV/MX path instead sums all represented E2M1 payload words.

\subsection{Policy ladder and distribution dependence}

The NV/NV investigation exposed \code{fast}, \code{universal}, and
\code{hao-l2} style policies. \code{Fast} used the most aggressive
shiftless approximation. \code{Universal} restored scale guards and a more
reliable row reference. \code{Hao-l2} retained full stabilization. Their
rough latency hierarchy at S4096/H24 was approximately 0.10, 0.11--0.13,
and 0.188\,ms, respectively, depending on the exact generation of the
kernel.

The important result was not one latency number. Fast NV/NV could have good
cosine on Gaussian scores and still generate non-finite or highly distorted
model outputs when its row-reference assumptions changed. Stabilization
fixed those cases but consumed the desired speedup. Fixed-scale sweeps on
the shiftless estimator did not reproduce the accuracy of the fully
stabilized E4M3 control.

\subsection{Why NV/NV was not retained}

NV/NV contributed three ideas that remain: code-directed approximation,
early partial P publication, and the need to evaluate speed and downstream
accuracy together. It was superseded for four reasons:
\begin{enumerate}
  \item an E4M3 P scale without a robust global factor can round
  low-amplitude N32 blocks to zero; applying the factor correctly fixes the
  format-level underflow but restores work on the exposed P path;
  \item independent block-16 scale work does not align as naturally with the
  N32 software producer;
  \item the robust stabilized path was too slow, while the fastest shiftless
  path was distribution-sensitive;
  \item the HAO-derived 512-column layout and MXFP4 scale handoff provided a
  cleaner way to retain small P blocks without adding another score bank.
\end{enumerate}

The fixed-schedule format table remains in the main paper because it is a useful
control: NV/NV can outperform NV/MX on a specific synthetic metric. The
format-range and downstream experiments explain why that isolated result is
not sufficient to select the production route.

\subsection{B300 D64 sweep and range failure}
\label{app:nvnv-d64}

The D64 specialization provides a wider test of this distinction. Its B300
NV/MX route combines fused TMEM load/reduction with eight native-EX2 pairs and
eight affine pairs per score quarter. Every NV/MX output in the 24-shape
matrix is finite. The raw shiftless NV/NV control is often locally more
accurate because its E4M3 scale can fit a benign block more closely, but it is
finite on only \BThreeDsixtyfourNVNVFiniteCases\ of the 24 tested cases.

\begin{table}[htbp]
\centering
\caption{Selected B300 D64 format diagnostics. NV/MX is the retained finite
route. NV/NV timings are shown only to explain the rejected control and must
not be treated as production performance when the status is non-finite.}
\label{tab:b300-d64-diagnostic}
\scriptsize
\begin{adjustbox}{max width=\textwidth}
\begin{tabular}{rrrrrrrrrl}
\toprule
$H$ & $S$ & \multicolumn{4}{c}{NV/MX} & \multicolumn{3}{c}{raw NV/NV} & Status \\
\cmidrule(lr){3-6}\cmidrule(lr){7-9}
 & & ms & TFLOP/s & Cosine & Rel.-$L_2$ & ms & Cosine & Rel.-$L_2$ & \\
\midrule
12 & 4096 & 0.066528 & 775 & 0.955818 & 0.295719 & 0.070464 & 0.962281 & 0.278304 & finite \\
24 & 4096 & 0.093248 & 1105 & 0.955551 & 0.296653 & 0.099296 & 0.962063 & 0.279040 & finite \\
24 & 32768 & 4.601808 & 1434 & 0.956089 & 0.295010 & 4.842496 & 0.962065 & 0.279155 & finite \\
32 & 2048 & 0.039872 & 862 & 0.954553 & 0.299511 & 0.039968 & 0.961772 & 0.280428 & finite \\
32 & 32768 & 6.115360 & 1438 & 0.957303 & 0.291325 & 6.465520 & -- & -- & non-finite \\
64 & 1024 & 0.025568 & 672 & 0.952035 & 0.306387 & 0.025632 & 0.960558 & 0.286304 & finite \\
64 & 4096 & 0.207840 & 1323 & 0.955838 & 0.295956 & 0.218176 & 0.962128 & 0.278683 & finite \\
64 & 32768 & 12.223360 & 1439 & 0.957248 & 0.291501 & 12.919808 & 0.962800 & 0.276050 & finite \\
\bottomrule
\end{tabular}
\end{adjustbox}
\end{table}
\FloatBarrier

Across S4096-and-longer rows, retained NV/MX is
\BThreeDsixtyfourMinSpeedup--\BThreeDsixtyfourMaxSpeedup$\times$ faster than
the matched B300 BF16 kernel. Density two wins or ties density one on
\BThreeDsixtyfourDensityWins\ of 24 shapes. H32/S2048 and H64/S1024 need
136-CTA and 128-CTA grid caps, respectively, because their short logical job
counts interact poorly with the full 148-worker persistent grid. Other shapes
retain 148 CTAs. Every promoted build uses 128 registers, one barrier, and no
local-memory spills.

The decisive failure occurs at H32/S32768. Seed 20260802 produces exactly 64
non-finite raw NV/NV outputs while BF16 and NV/MX remain finite; another seed
passes. Saturating the E4M3 scale encoder repairs the failing distribution at
\BThreeDsixtyfourBoundedNVNVTime\,ms and
\BThreeDsixtyfourBoundedNVNVSpeedup$\times$ BF16, with 0.962921 cosine and
0.275742 relative-$L_2$. This bounded mode costs roughly 2--3\%. MXFP4's E8M0
scale is not finer than E4M3, but its exponent range represents tiny
probability blocks without a separately materialized row shift. Raw NV/NV can
therefore win a local error metric when it remains in range; NV/MX is the
retained low-latency route because it remains finite across the matrix.

\subsection{Measured downstream failure mechanism}
\label{app:nvnv-failure}

The expanded provider matrix isolates the failure. Native HAO and the TK
fixed-schedule control receive the same NVFP4-quantized Q, K, and V tensors.
HAO first subtracts a row maximum. The TK control instead forms a shiftless
block scale from
\begin{equation}
  s_B^{\mathrm{shiftless}}
  = \frac{1}{6}\max_{j\in B}\exp(z_j),
\end{equation}
then encodes it in E4M3. Model scores make this quantity exceed the finite
E4M3 maximum of 448 in every tested workload. This control has no
satfinite guard, so its out-of-range scale conversion contaminates the
dependent P and denominator path and produces non-finite context rows.

With row-max stabilization,
\begin{equation}
  s_B^{\mathrm{stable}}
  = \frac{1}{6}\max_{j\in B}\exp(z_j-m),
  \qquad m=\max_j z_j,
\end{equation}
so $0\le s_B^{\mathrm{stable}}\le 1/6$. Table~\ref{tab:nvnv-failure}
reports the first-sample diagnostic. The non-finite count is accumulated
over all attention layers evaluated for that sample. Small stabilized blocks
can still underflow, but they contain negligible probability mass relative
to the row anchor; HAO remains finite on every full fixed-input evaluation.

\begin{table}[htbp]
\centering
\caption{Shiftless TK NV/NV failure on model activations. Overflow is the
fraction of N32 P scales above E4M3's maximum before encoding.}
\label{tab:nvnv-failure}
\scriptsize
\begin{adjustbox}{max width=\textwidth}
\begin{tabular}{lrrrrrr}
\toprule
Task & Failed sample & Non-finite rows & Shiftless overflow (\%)
& Shiftless max & Stable overflow (\%) & Stable max \\
\midrule
ViT S256 & 1 & 2,560 & 8.146 & 1.828e+37 & 0.000 & 0.166667 \\
ViT S1024 & 1 & 127,950 & 8.137 & 5.332e+37 & 0.000 & 0.166667 \\
ViT S4096 & 1 & 539,442 & 7.511 & 5.669e+37 & 0.000 & 0.166667 \\
BERT MLM S256 & 1 & 167 & 1.721 & 5.279e+04 & 0.000 & 0.166667 \\
BERT MLM S512 & 1 & 49,613 & 1.063 & 5.655e+04 & 0.000 & 0.166667 \\
BERT SST-2 S256 & 1 & 21,717 & 1.465 & 4.493e+03 & 0.000 & 0.166667 \\
\bottomrule
\end{tabular}
\end{adjustbox}
\end{table}

\section{Rejected Kernel Directions}
\label{app:rejections}

\subsection{SM103 launch and EX2 controls}
\label{app:sm103-tuning}

Table~\ref{tab:b300-tuning} records the initial B300 compatibility and
native-EX2 density sweep. It is kept outside the main result table because it
is a tuning control, not a set of promoted policies. The 152-worker binary
creates a partial second wave on the tested 148-SM SKU; the corrected rows use
148 workers and vary EX2 density while holding the numerical contract fixed.

\begin{table}[htbp]
\centering
\caption{Measured TK NV/MX tuning points on B300. Every row reports speed and
error from the same output.}
\label{tab:b300-tuning}
\scriptsize
\begin{adjustbox}{max width=\textwidth}
\begin{tabular}{lrrrrrrr}
\toprule
Variant & Grid & EX2 density & Time (ms) & TFLOP/s & Cosine & Rel.-$L_2$ & RMSE \\
\midrule
B300 compatibility & 152 & 0 & 0.123872 & 1664 & 0.943964 & 0.335825 & 0.008716 \\
B300 density 0 & 148 & 0 & 0.093216 & 2212 & 0.943964 & 0.335825 & 0.008716 \\
B300 density 1 & 148 & 1 & 0.097248 & 2120 & 0.947037 & 0.325258 & 0.008442 \\
B300 density 2 & 148 & 2 & 0.097280 & 2119 & 0.950069 & 0.314812 & 0.008171 \\
B300 density 4 & 148 & 4 & 0.111584 & 1848 & 0.956047 & 0.294037 & 0.007632 \\
\bottomrule
\end{tabular}
\end{adjustbox}
\end{table}

\subsection{Kernel directions}

Table~\ref{tab:rejections} condenses the major negative experiments. These
results constrain the interpretation of the retained method; compile-time
probes remain default-off.

\begin{table}[htbp]
\centering
\caption{Major rejected directions. A timing tie is not promoted when it
adds synchronization, storage, or numerical risk.}
\label{tab:rejections}
\begin{tabularx}{\textwidth}{p{.19\textwidth}X X}
\toprule
Direction & Intended benefit & Observed failure mechanism \\
\midrule
Half-tile QK/PV & Larger tensor work and easier overlap & Delayed first
publication and increased live tensor-memory pressure; did not beat N32
production with K64 consumption. \\

Deeper dynamic scheduler & Exploit QK running one logical step ahead & Polls,
proxy signals, and policy branches added control work without creating a
legal score destination. \\

Full or QK-only two-CTA & Accelerate QK and increase occupancy & Cluster-wide
readiness and scale lifetime coordination overwhelmed QK savings; QK-only
did not remove single-CTA P/PV ownership. \\

Extra barriers or offload WG & Remove full-CTA rendezvous & Duplicate score
loads and handoff mailboxes cost more than the hidden work; concurrent TMEM
writes produced invalid output. \\

Alternate TMEM layouts & Add a second useful score/P slot & Two 128-column
scores plus two 128-column outputs already consume 512 columns. Scale
compression freed fragments, not another legal 128-column bank. \\

BF16/FP16 partial accumulator & Halve output columns & Local scaled-FP4 tensor
instructions accumulate into FP32 TMEM; casting between issues did not change
the accumulator contract. \\

\bottomrule
\end{tabularx}
\end{table}

\begin{table}[htbp]
\centering
\caption{Major rejected directions (continued).}
\begin{tabularx}{\textwidth}{p{.19\textwidth}X X}
\toprule
Direction & Intended benefit & Observed failure mechanism \\
\midrule

Initial NVFP4 projection path & Extend FP4 tensor throughput across the learned
attention projections & Isolated D128 attribution found much larger projection
error than E4M3 around an otherwise faithful attention core.  We dropped that
implementation, not the format: later distributed experiments use NVFP4
projections as the higher-throughput arm.  This result says nothing about
NVFP4 Q/K inside attention. \\

Raw FP4 or coarse 2-D scales & Remove scale pages & Raw E2M1 loses the
four-times-class block-scaled primitive. A single 32$\times$32 scale cannot
be applied after a reduction whose block product scales vary with K. \\

Direct code classifier & Eliminate packed conversion & Threshold trees,
integer conversion, LUT access, LOP3, and PRMT packing generated more SASS
than packed FFMA2 plus native F2FP. \\

Intermediate NV/MX policy & Add an anchor without the correction warpgroup &
At S4096/H24 it measured 0.094560\,ms, slower than \code{fast}, while its
0.356700 relative-$L_2$ was worse than both \code{fast} and \code{accurate}.
Its long-ViT agreement only tied \code{accurate}, leaving no Pareto value. \\

Quadratic/cubic throughput path & Improve code fit & A Q0 quadratic raised
static FFMA2 count from 128 to 160, measured 0.097888\,ms, and reduced cosine
on its test. \\

Sampled max/denominator & Shorten P preparation & Eight-of-32 samples saved
less than 0.5\us\ with substantial error; fewer samples were unstable. \\

Structured sparse PV & Increase tensor throughput & Blackwell's sparse FP4
path uses logical K128, losing the early K64 handoff; value-aware selection
added too many instructions. \\

Tail interleaving & Pull Q3 work under Q2/PV latency & Added 1.4--3.2\us;
the contiguous Q2-then-Q3 schedule was locally better. \\

Streaming denominator & Hide exact denominator reduction & Preserved
output exactly but slowed the clean fast build by 2.11\% and 1.83\%.
\\
\bottomrule
\end{tabularx}
\end{table}

The raw cases for the retired intermediate policy remain in the unified
suite for provenance, but it is excluded from the published CSV, plots, and
policy tables.

\subsection{Quarter-local speed-of-light experiment}

On the historical 0.102400\,ms NV/NV path, replacing selected quarter
transforms with raw packing measured:
\begin{center}
\begin{tabular}{lrr}
\toprule
Transform removed & Time (ms) & Gain \\
\midrule
Q0 only & 0.104736 & none \\
Q1 only & 0.102400 & none \\
Q0+Q1 & 0.099328 & 4.064\us \\
Q2+Q3 & 0.096416 & 7.008\us \\
all quarters & 0.090112 & 12.288\us \\
\bottomrule
\end{tabular}
\end{center}
The pair behavior is the important finding: accelerating one N32 producer
does not advance a K64 tensor command. The 12.288\us\ total is not the
current NV/MX ceiling; Section~\ref{sec:current-headroom} reports the final
2.336\us\ raw-score-pack gap.

\subsection{Scale-footprint experiments}

Several probes attempted to reduce scale storage by sharing scales across
32$\times$32 blocks, folding adjacent IDs, or moving source scales through
shared memory. Compact source storage is useful, but \tcgen\ still consumes
an instruction-facing scale layout. A smaller source tensor therefore does
not automatically create free tensor-memory columns. The retained folded
K64 Q/K representation reduces loads and arithmetic; it does not claim a
new 128-column score slot.

\subsection{Joint affine-route search}
\label{app:joint-affine}

We also optimized the affine E2M1 coefficients across all non-guard layers at
once. Exact, equal-cost binaries were screened with persistent model workers;
the optimizer combined favorable single-layer changes and then sampled full
categorical layer maps. This produced large apparent fixed-input gains, but fresh
model processes rejected every proposed production change.

For Wan14B, the best regularized map changed four layers and reduced reused-
worker mean relative-$L_2$ from 0.398533 to 0.392793 over four prompts. On a
fresh city run it instead increased relative-$L_2$ from 0.432932 to 0.453884.
A fresh forest run also became non-finite in guarded layer 38, but that event
was later traced to the E8M0 denominator-underflow bug described with the main
Wan results; it is not evidence for or against the affine map.
For Wan1.3B, a single layer-9 change reduced reused-worker mean
relative-$L_2$ from 0.299884 to 0.290924. Fresh coastal and snow runs both
regressed: 0.247742 to 0.258156 and 0.230262 to 0.233996, respectively.

The route constants do not change kernel latency; the rejection is purely
numerical. Reused-pipeline ordering can move the final diffusion latent by
more than the candidate margin and make a small ranking unstable. A
statistically sound joint optimizer would therefore need to score each route
across several fresh model processes and held-out prompts. That cost is not
justified here because it cannot improve throughput and the observed numerical
margins are below process-to-process variation. We keep the joint optimizer as
a diagnostic and require fresh-process, unseen-prompt validation for any
future calibration change. The production manifest retains only the global
$(1.60,0.95)$ pair. Full search and control artifacts are in
\path{results/fp4_fa4_wan_joint_20260806}.

\section{Accuracy-Matched FP8 Control}
\label{app:accuracy-matched-control}

The principal NV/MX result is a speed--accuracy trade-off, so a separate
control asks what happens when the TK route is required to match HAO's
reported NV/FP8 cosine. Table~\ref{tab:accuracy-matched-control} uses the
B1/S32768/H24/D128 B300 record. The HAO rows are reconstructed from published
TFLOP/s and are cross-run context \cite{hao2026fp4flashattention}; HAO does
not publish relative-$L_2$.

\begin{table}[ht]
\centering
\caption{Accuracy-matched B300 control. Superscript p marks HAO-published
cross-run values.}
\label{tab:accuracy-matched-control}
\scriptsize
\begin{adjustbox}{max width=\columnwidth}
\begin{tabular}{lrrrr}
\toprule
Provider & Time (ms) & TFLOP/s & Cosine & Relative-$L_2$ \\
\midrule
TK NV/MX fast & 4.481 & 2945 & 0.9429 & 0.3389 \\
TK NV/FP8 optimized & 7.584 & 1740 & 0.9573 & 0.2913 \\
TK NV/FP8 exact & 8.854 & 1490 & 0.9897 & 0.1433 \\
HAO NV/FP8\textsuperscript{p} & 4.929 & 2677 & 0.9899 & -- \\
HAO BF16\textsuperscript{p} & 8.607 & 1533 & 1.0000 & -- \\
\bottomrule
\end{tabular}
\end{adjustbox}
\end{table}

The exact TK NV/FP8 route reaches 0.9897 cosine, effectively matching HAO's
published 0.9899 at the displayed precision. It takes 8.854 ms and reaches
1490 TFLOP/s, compared with HAO's 4.929 ms and 2677 TFLOP/s. It also slightly
trails HAO's published BF16 throughput. The intermediate optimized TK NV/FP8
route improves speed but reaches only 0.9573 cosine. This experiment separates
two conclusions: the HAO-derived pipeline is structurally viable, while the
large speed of our retained NV/MX mode comes from making the exposed P path
cheaper and accepting lower operator fidelity. Matching HAO's accuracy with
the current exact TK FP8 path gives back that advantage.

\section{Complete Fixed-Schedule Format Matrix}
\label{app:format-matrix}

The main text reports the headline fixed-schedule control because the
retained-policy frontier is the principal result. Table~\ref{tab:all-formats}
provides the corresponding control at every unified shape. The four TK rows
at each shape use one transform and scheduling budget while changing only the
QK and P/V scale formats. Native HAO NV/NV and BF16 are included as external
controls. Timing and all three error metrics come from the same deterministic
shape record and use the same canonical BF16 reference as the main tables.
This matrix concerns Q/K/P/V formats inside the forward-only attention kernel;
it neither includes learned QKV/O projections nor implies a pure-FP4
end-to-end training route.

\begin{table}[htbp]
\centering
\caption{Complete fixed-schedule format matrix. Each row reports latency and output
error together; no timing is paired with accuracy from another run.}
\label{tab:all-formats}
\scriptsize
\begin{tabular}{llrrrrr}
\toprule
Shape & Provider & Time (ms) & Speedup & Cosine & Relative-$L_2$ & RMSE \\
\midrule
B1/S256/H16 & TK NV/NV fixed schedule & 0.010560 & 1.165$\times$ & 0.952388 & 0.320599 & 0.032716 \\
 & TK MX/NV fixed schedule & 0.010528 & 1.169$\times$ & 0.948314 & 0.320161 & 0.032672 \\
 & TK NV/MX fixed schedule & 0.010880 & 1.131$\times$ & 0.929613 & 0.373055 & 0.038069 \\
 & TK MX/MX fixed schedule & 0.010528 & 1.169$\times$ & 0.924689 & 0.380760 & 0.038856 \\
 & HAO NV/NV & 0.014336 & 0.858$\times$ & 0.982173 & 0.188921 & 0.019279 \\
 & HAO BF16 & 0.012304 & 1.000$\times$ & 1.000000 & 0.000000 & 0.000000 \\
B1/S1024/H24 & TK NV/NV fixed schedule & 0.018240 & 1.245$\times$ & 0.960499 & 0.285533 & 0.014735 \\
 & TK MX/NV fixed schedule & 0.018176 & 1.249$\times$ & 0.953730 & 0.301114 & 0.015539 \\
 & TK NV/MX fixed schedule & 0.018432 & 1.232$\times$ & 0.946337 & 0.323387 & 0.016688 \\
 & TK MX/MX fixed schedule & 0.018432 & 1.232$\times$ & 0.937614 & 0.350956 & 0.018111 \\
 & HAO NV/NV & 0.033152 & 0.685$\times$ & 0.981844 & 0.190610 & 0.009836 \\
 & HAO BF16 & 0.022704 & 1.000$\times$ & 1.000000 & 0.000000 & 0.000000 \\
B1/S2048/H24 & TK NV/NV fixed schedule & 0.041248 & 1.492$\times$ & 0.961651 & 0.281034 & 0.010218 \\
 & TK MX/NV fixed schedule & 0.041632 & 1.478$\times$ & 0.954486 & 0.298556 & 0.010855 \\
 & TK NV/MX fixed schedule & 0.042080 & 1.463$\times$ & 0.948649 & 0.317135 & 0.011531 \\
 & TK MX/MX fixed schedule & 0.041280 & 1.491$\times$ & 0.939444 & 0.347612 & 0.012639 \\
 & HAO NV/NV & 0.072112 & 0.854$\times$ & 0.981680 & 0.191501 & 0.006963 \\
 & HAO BF16 & 0.061552 & 1.000$\times$ & 1.000000 & 0.000000 & 0.000000 \\
B1/S4096/H24 & TK NV/NV fixed schedule & 0.104000 & 1.585$\times$ & 0.962721 & 0.276316 & 0.007167 \\
 & TK MX/NV fixed schedule & 0.102688 & 1.605$\times$ & 0.955444 & 0.295344 & 0.007660 \\
 & TK NV/MX fixed schedule & 0.104032 & 1.584$\times$ & 0.950622 & 0.311853 & 0.008088 \\
 & TK MX/MX fixed schedule & 0.102976 & 1.600$\times$ & 0.941142 & 0.343985 & 0.008922 \\
 & HAO NV/NV & 0.192512 & 0.856$\times$ & 0.981894 & 0.190474 & 0.004940 \\
 & HAO BF16 & 0.164800 & 1.000$\times$ & 1.000000 & 0.000000 & 0.000000 \\
B1/S4096/H64 & TK NV/NV fixed schedule & 0.227328 & 1.631$\times$ & 0.962449 & 0.277441 & 0.007171 \\
 & TK MX/NV fixed schedule & 0.235616 & 1.573$\times$ & 0.955133 & 0.296365 & 0.007660 \\
 & TK NV/MX fixed schedule & 0.232128 & 1.597$\times$ & 0.950186 & 0.313010 & 0.008090 \\
 & TK MX/MX fixed schedule & 0.234112 & 1.583$\times$ & 0.940680 & 0.345061 & 0.008918 \\
 & HAO NV/NV & 0.438192 & 0.846$\times$ & 0.981743 & 0.191187 & 0.004941 \\
 & HAO BF16 & 0.370688 & 1.000$\times$ & 1.000000 & 0.000000 & 0.000000 \\
B1/S8192/H64 & TK NV/NV fixed schedule & 0.861792 & 1.870$\times$ & 0.962801 & 0.276083 & 0.005045 \\
 & TK MX/NV fixed schedule & 0.905248 & 1.780$\times$ & 0.955295 & 0.295814 & 0.005405 \\
 & TK NV/MX fixed schedule & 0.888832 & 1.813$\times$ & 0.950828 & 0.311299 & 0.005688 \\
 & TK MX/MX fixed schedule & 0.903488 & 1.784$\times$ & 0.941102 & 0.344364 & 0.006292 \\
 & HAO NV/NV & 1.693696 & 0.951$\times$ & 0.981700 & 0.191467 & 0.003499 \\
 & HAO BF16 & 1.611488 & 1.000$\times$ & 1.000000 & 0.000000 & 0.000000 \\
\bottomrule
\end{tabular}
\end{table}
\FloatBarrier

\section{Reproduction and Compile-Time Contract}
\label{app:reproduction}

The public code release is available at \projectrepo. It contains the TK
forward and backward kernels, the CuTe-DSL comparison and prototype kernels,
experiment configurations, committed evidence, and the scripts used to build
this paper.

\subsection{Canonical command graph}

The release verifier authenticates the committed source and historical
snapshot bytes. Fresh measurements can be generated from the repository root with
\path{tools/plan_fa4_measurements.py}. The planner selects the historical
snapshot under \path{reproduction/snapshots/forward_cfc06dad} for the
noncausal study and the root causal source for the training study. It checks
supplied causal artifacts and external assets, then emits a fresh build and run
graph; it does not download data, run CUDA, or submit a job. Newly built
extensions are checked against generated bundle manifests or recorded by
content digest when the generated command executes.

List the complete graph with:
\begin{lstlisting}[language=bash]
python3 tools/plan_fa4_measurements.py list
\end{lstlisting}
For example, this validates and prints the noncausal forward graph:
\begin{lstlisting}[language=bash]
python3 tools/plan_fa4_measurements.py print \
  --family noncausal-forward \
  --python /absolute/path/python3 \
  --output-root /absolute/path/new-results \
  --noncausal-build-root /absolute/path/new-build \
  --cuda-home /absolute/path/cuda-13.0 \
  --cutlass-dsl-root /absolute/path/cutlass-dsl
\end{lstlisting}
The output and build roots must be new or empty. A blocked node is printed as a
comment and makes the planner exit with status 2; missing evidence is never
replaced by a guessed command. The generated noncausal suite preallocates
outputs, cycles provider order, and records every timing window.

\subsection{Fixed-input model evaluations}

ViT, BERT masked-language-model, and SST-2 replacement measurements use the
\code{downstream} family. Each generated process runs exactly one task with an
explicit extension root and one authenticated model/dataset pair:
\begin{lstlisting}[language=bash]
python3 tools/plan_fa4_measurements.py print \
  --family downstream \
  --python /absolute/path/python3 \
  --output-root /absolute/path/new-results \
  --noncausal-build-root /absolute/path/new-build \
  --cuda-home /absolute/path/cuda-13.0 \
  --cutlass-dsl-root /absolute/path/cutlass-dsl \
  --external-assets-manifest /absolute/path/assets.json
\end{lstlisting}
The asset manifest binds immutable revisions and every local file by byte count
and SHA256. These commands produce new, fully pinned measurements. The
historical paper rows remain receipt-backed because their original external
asset revisions were not recorded.

The \code{vit-mae} and \code{wan} families use the same asset contract. The
ViT-MAE manifest must identify the recorded 100 COCO validation images. Wan
keeps the logical model identifier separate from the authenticated local model
snapshot and builds the fast and accurate policy bundles before evaluation. The
paired HAO-BF16/TK comparison is available; HAO low-precision Wan controls remain
blocked because their exact extension build identity was not preserved.

The committed Wan table is a historical artifact. Some calibration and
provider JSON inputs needed to regenerate it are absent, so the release does
not advertise \path{build_tables.py} as a complete rebuild. The retained
receipts support the reported values, while a fresh \code{wan} run produces
replacement evidence under the current authenticated protocol.

\subsection{Offline artifacts and external inputs}

Tables, figures, and manuscript outputs supported by committed inputs are
rebuilt without network access using:
\begin{lstlisting}[language=bash]
python3 tools/reproduce_fa4_paper.py --run --offline all
\end{lstlisting}
\path{docs/fa4_measurement_reproduction.md} defines the external-asset and
artifact-manifest schemas. It also records which historical measurements are
receipt-only and which can be replaced by a fully authenticated fresh run.

\subsection{B300 aggregate reconstruction}

The source release contains the committed aggregate B300 summary used by the
paper, but not the larger raw cluster-capture archive. Regenerate the LaTeX
tables from that committed summary with:
\begin{lstlisting}[language=bash]
python3 results/fp4_fa4_b300_tuning_20260802/\
build_summary.py --from-summary
\end{lstlisting}
This route performs no network access and validates the summary schema before
rendering. Recomputing the aggregate from raw captures remains blocked until a
checksummed raw-capture bundle is published separately. Once available,
restore it beneath the B300 result directory and run
\code{build\_summary.py} without \code{--from-summary}. Cluster submission
files, scheduler identifiers, private storage locations, and secret wiring are
operational metadata rather than scientific inputs and are excluded from the
public reproduction path.

\subsection{Required operand contract}

Both retained policies require folded K64 Q/K scales selected by the MSE
policy:
\begin{lstlisting}[language=bash]
--nv-qk-fold-k64-scales both \
--nv-qk-fold-scale-select mse
\end{lstlisting}
\code{Accurate} also requires the same fixed 32-row permutation for K and V.
Pairing a folded-scale binary with ordinary block-16 Q/K scales can produce
plausible timing and invalid accuracy; the suite couples the binary and
operand arguments.

\subsection{Retained compile-time policy}

\code{Fast} uses a 12-stage K/V ring, all-affine mode 23, four-word
represented denominators, paired scale reuse, wide three-input maxima,
Q-scale preloading, early asynchronous P-scale handoff, and a 200/208/56
register split. \code{Accurate} uses 13 K/V stages, native EX2 samples, a
32-row anchor, and correction-warpgroup normalization.

Historical NV/NV, intermediate NV/MX, sparse, sampled-denominator,
direct-code, quadratic, two-CTA, alternate-owner, and interleaving probes
remain compile-gated and default-off.

\section{Terminology}

\begin{description}
  \item[Direct-P] The retained forward-inference method: NVFP4 Q/K,
  MXFP4 P/V, direct log-score-to-E2M1 probability conversion, and
  normalization from the represented probability consumed by PV.
  \item[Quantized causal backward] The retained training method: reconstruct
  P from the saved quantized Q/K payload, scales, and LSE, then use
  range-appropriate, matrix-oriented FP8 operands for gradient products.
  \item[BF16 / FP8 / FP4] Bfloat16 and 8-/4-bit floating-point families.
  E$x$M$y$ denotes $x$ exponent bits and $y$ explicit fraction bits.
  \item[SM / CTA] A streaming multiprocessor is one GPU compute unit; a
  cooperative thread array is a CUDA thread block scheduled on an SM.
  \item[TMEM / TMA] Tensor memory is Blackwell's on-chip matrix-accumulator
  scratchpad.  The Tensor Memory Accelerator moves tiles between GPU memory
  spaces.
  \item[MMA] A tensor-core matrix multiply--accumulate operation.
  \item[FMA / SFU / EX2] Fused multiply--add; special-function unit; and the
  hardware base-two exponential instruction.
  \item[GQA / RoPE / LSE] Grouped-query attention; rotary positional
  embedding; and the per-row log-sum-exp softmax normalizer.
  \item[Quarter] One N32 fragment read from an N128 score tile. Four producer
  quarters feed two K64 scaled-FP4 PV commands.
  \item[Score bank] A 128-column FP32 TMEM region receiving QK and later
  hosting transient P/scale overlays.
  \item[Output bank] A permanent 128-column FP32 PV accumulator for one query
  stage.
  \item[Represented denominator] A normalizer summed from the E2M1 payload
  and decoded block scale actually consumed by PV.
  \item[Early publication] Signaling a legal first K64 payload and scale pair
  before tail P construction completes.
  \item[Speed-of-light diagnostic] An intentionally incorrect or
  semantically altered kernel that removes work to bound latency.
  \item[Full FP4 attention] Q, K, P, and V inside the attention kernel use
  E2M1 payloads with block scales. NVFP4 QK plus FP8 PV is a control, not full
  FP4 attention.  The term makes no claim about learned QKV/O projections or
  an end-to-end training recipe.
\end{description}

\section{What We Tried in Causal Training---and Why We Did Not Keep It}
\label{app:causal-design-history}

The retained method in Section~\ref{sec:causal-training} is the end of a much
larger search.  This appendix gives the useful lessons without requiring the
reader to follow the order in which kernels were written.  Internal version
names appear only here and in the provenance appendix.

\subsection{Projection precision and attention precision are different}

The first important separation is between learned projection layers and the
attention kernel.  QKV and output projections multiply model activations by
learned weights; attention then consumes the resulting Q, K, and V tensors.
Using FP4 successfully inside attention does not imply that learned weights
and activations can use the same format.

In an isolated 8B transformer block, E4M3 QKV projection had approximately
0.0375 relative-$L_2$ error, compared with 0.1459 for NVFP4 QKV.  When the
decoded low-precision operands were passed through exact bfloat16 attention,
the attention-output error was already 0.3614.  The native attention kernel added only
0.0216 relative-$L_2$ on those same represented operands.  In other words,
most error entered before the attention kernel.  This is why the main
experiments treat projection precision as a separate variable: E4M3 is the
numerical control, while NVFP4 is the higher-throughput arm.  This isolated
block does not by itself choose a final projection format.

An earlier fixed-head scaling rule also failed immediately on the 8B shape.
Replacing it with two-dimensional row-by-K16 Q/K scales reduced pre-clipping
gradient norms from roughly 20,000 to 161--184 in the short gate and restored
bfloat16-like behavior.  This was a representation fix, not a change to the
matrix schedule.

\subsection{Two independent backward scale failures}

Backward initially produced finite tensors that were nevertheless wrong.  Two
independent factors of 256 were involved:

\begin{enumerate}
  \item score reconstruction consumes a lifted statistic
  $\ell_i=8-L_i\log_2(e)$, where $L_i$ is the saved natural-log LSE.  Omitting
  $+8$ reconstructs the probability at $1/256$ of its intended scale;
  \item the gradient-output epilogue has its own $1/256$ correction.  It is a
  separate matrix-layout convention and must not be folded into the LSE fix.
\end{enumerate}

After those corrections, a checkpoint diagnostic exposed a different problem:
E4M3 rounded about 97.1\% of dO values to zero.  E5M2 has fewer fraction bits
but a wider exponent range; publishing dO in E5M2 reduced the observed zero
fraction to about 14.4\%.  The E5M2 publisher was only 0.7--0.9\% slower than
the E4M3 publisher in the matched gate.  This is why the final method uses two
different FP8 formats rather than treating all 8-bit values as interchangeable.

\subsection{How the native backward schedule evolved}

Table~\ref{tab:causal-version-map} maps internal development labels to the
idea each one tested.  The labels are not public method names.

Some measurements apply only to predecessor schedules.  In the v501 B2 owner4
route, dK and dV had unique writers, so only dQ required clearing at the exact
B2/S4096 shape.  A bounded profile of that same predecessor measured 21\% warp
occupancy, 30\% tensor activity, and 9\% external-memory activity despite 95\%
SM activity.  The retained v509 entry point always clears its outputs; no
equivalent performance profile exists for that final binary.  Reuse of rounded
P, dual-layout dS publication, and early tensor-memory release remain useful
design ideas, but predecessor counters are not evidence for final-route speed.

\begin{table}[htbp]
\centering
\caption{Internal backward version map.}
\label{tab:causal-version-map}
\small
\begin{tabularx}{\textwidth}{l p{.29\textwidth}X}
\toprule
Label & Main purpose & Outcome \\
\midrule
v416 & D64 native owner schedule & Like-for-like parity with the generated
native-exponential reference; used in early 1.2B integration. \\
v454/v482 & D128 B1/B2 ownership, rounded-P reuse, and early tensor-memory
release & 1.21--1.24$\times$ faster than the matched generated reference. \\
v501 & Corrected LSE lift, shape-specific clearing, and represented E4M3
gradient operands & Finite short-run systems prototype and basis of the
historical 8B bracket. \\
v503 & Common-row MXFP4 V approximation in backward & Faster than the first
MX attempt and tied end to end; the complete recipe failed its observed
distributed numerical gate, but the consumer was not isolated as the cause. \\
v506/v507 & Direct shared-MX producer and exact four-anchor consumer &
Numerically useful controls, but too slow for the production gate. \\
v509 & Exact forward NVFP4 score reconstruction with E5M2 dO & Retained
quantized causal-backward implementation; exact-batch B1/B2/B4 binaries are
validated for the Llama-style D128 shape. \\
\bottomrule
\end{tabularx}
\end{table}
\FloatBarrier

\subsection{Historical isolated backward timings}

The longest matched bfloat16 sequence sweep used a generated CuTe kernel, not
the retained implementation.  It nevertheless records how that earlier
low-precision path scaled from 512 to 16,384 tokens.

\begin{table}[htbp]
\centering
\caption{Historical isolated D64 causal backward on GB200.  The low-precision
route is a generated CuTe kernel; both columns include required output clears.}
\label{tab:d64-bf16-sweep}
\small
\begin{tabular}{rrrr}
\toprule
Sequence & Exact BF16 & Low precision & Speedup \\
\midrule
512   & 111.456 $\mu$s  & 102.176 $\mu$s  & 1.091$\times$ \\
1024  & 149.504 $\mu$s  & 138.560 $\mu$s  & 1.079$\times$ \\
2048  & 203.808 $\mu$s  & 189.632 $\mu$s  & 1.075$\times$ \\
4096  & 347.104 $\mu$s  & 319.808 $\mu$s  & 1.085$\times$ \\
8192  & 879.168 $\mu$s  & 770.848 $\mu$s  & 1.141$\times$ \\
16384 & 2765.984 $\mu$s & 2621.376 $\mu$s & 1.055$\times$ \\
\bottomrule
\end{tabular}
\end{table}
\FloatBarrier

At D128, the optimized native predecessor was compared with a generated
low-precision reference using the same represented operands.  This isolates
scheduling improvement; it is not a low-precision-versus-BF16 comparison.

\begin{table}[htbp]
\centering
\caption{Historical isolated D128 causal backward at S4096/Hq32/Hkv8 on
GB200.  The native column is the v454/v482 predecessor family.}
\label{tab:causal-backward-results}
\small
\begin{tabular}{lrrr}
\toprule
Shape & Generated reference & Native schedule & Speedup \\
\midrule
B1/D128 & 381.376 $\mu$s & 315.360 $\mu$s & 1.209$\times$ \\
B2/D128, rotation A & 620.032 $\mu$s & 514.272 $\mu$s & 1.206$\times$ \\
B2/D128, rotation B & 639.328 $\mu$s & 515.040 $\mu$s & 1.241$\times$ \\
\bottomrule
\end{tabular}
\end{table}
\FloatBarrier

\subsection{Why MXFP4 V in backward was not retained}

MXFP4 P/V is attractive in forward because the isolated D128/B2 core measures
259.072 $\mu$s, compared with 295.008 $\mu$s for FP8 P/V.  The full training
step, however, also needs V in a layout and format suitable for dP.  The first
MX route therefore published an MXFP4 V for forward and a second E4M3 V for
backward, consuming most of the isolated saving before backward began.

We tried to remove the second publication.  A shared-tile producer quantized
each D32-by-S32 V tile once and wrote both physical orientations.  An exact
four-anchor backward matched its represented oracle but took 581.632 $\mu$s,
versus 485.376 $\mu$s for the E4M3-V control.  The faster common-row consumer
reduced time but changed dQ/dK: cosines were about 0.989 and relative-$L_2$
about 0.145 against the E4M3-V route.  At whole-step scale, shared MX was only
0.194\% faster at the median and 0.102\% faster in sustained time---a tie, not
a useful throughput result.

Small S256 stochastic-rounding tests decoded MX values into E4M3 and therefore
served only as numerical proxies; they did not exercise a packed-MX kernel or
measure end-to-end time.  Those proxies and larger-batch checks did not change
the final choice.  The exact MX consumer was too slow, while the complete
approximate-consumer recipe failed the observed distributed numerical gate;
the failure was not isolated to that consumer.  The retained training path
therefore keeps E4M3 V views for backward and selects FP8 P/V in forward.

\subsection{Historical local speed brackets}

The early end-to-end brackets were valuable engineering checks, but they used
older learned-projection formats and cut-cross-entropy (CCE), so they are not
timings of the current dense-cross-entropy recipe.  Table
\ref{tab:historical-local-brackets} keeps them for context.

\begin{table}[htbp]
\centering
\caption{Historical saturated single-GPU brackets.  Each speedup is valid
within its row group but must not be transferred to the final training recipe.}
\label{tab:historical-local-brackets}
\small
\begin{tabular}{llrr}
\toprule
Model/shape & Attention route & Update time & Versus BF16 \\
\midrule
1.2B, B16/S4096 & BF16 FA4 & 673.396 ms & 1.000$\times$ \\
 & NVFP4-QK + FP8-PV & 615.682 ms & 1.094$\times$ \\
 & NVFP4-QK + MXFP4-PV & 614.842 ms & 1.095$\times$ \\
\midrule
8B, B2/S4096 & BF16 FA4 & 489.821 ms & 1.000$\times$ \\
 & NVFP4-QK + FP8-PV & 434.014 ms & 1.129$\times$ \\
 & NVFP4-QK + dual-published MXFP4-PV & 435.992 ms & 1.124$\times$ \\
\bottomrule
\end{tabular}
\end{table}
\FloatBarrier

The 8B FP8 and dual-published MX routes used the same backward.  Their backward
medians differed by only 0.084 ms, while the complete MX update was 1.978 ms
slower.  This explains the apparent paradox: an isolated MX forward kernel can
be materially faster even though its roughly one-millisecond saving per
32-layer model is small enough to be erased by publication and ordinary
step-level variation.

\subsection{Historical training context}

\begin{figure}[H]
\centering
\reportplot[0.27\textheight]{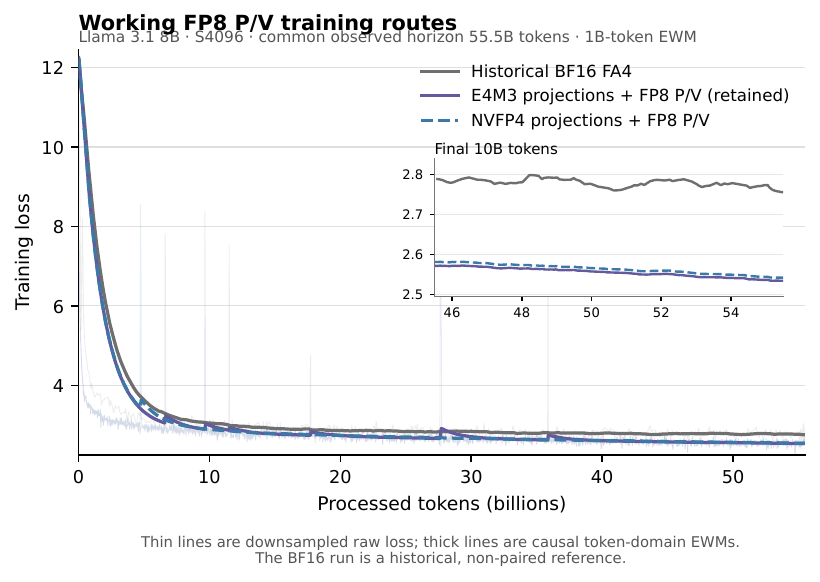}
\caption{Historical 8-billion-parameter training context through the common
55.5-billion-token horizon.  Thin lines show downsampled training loss and
thick lines show a causal exponential moving average with a one-billion-token
half-life.  The FP8-P/V arms use 64-way replicated data parallelism with
global batch 64.  The bfloat16 reference uses 32-way fully sharded data
parallelism with global batch 32; it is not a paired control.}
\label{fig:llama8b-training-curves}
\end{figure}

The September 1 snapshot contains two FP8-P/V trajectories that remain
non-divergent through a common 55.5-billion-token horizon.  Figure
\ref{fig:llama8b-training-curves} includes an older bfloat16 causal-FA4 run as
a sanity reference.  At that horizon, the one-billion-token exponential
moving-average losses are 2.534 for E4M3 projections with FP8 P/V, 2.542 for
NVFP4 projections with FP8 P/V, and 2.756 for the historical bfloat16 run.
These coordinates cannot be interpreted as relative model quality: the
bfloat16 run used a different topology, global batch, sample order, and
history-sampling density.

\subsection{What the distributed failures established}

The first long distributed MX run combined learned NVFP4 projections, CCE,
and the common-row MX backward, so its failure did not identify a single
cause.  At a matched checkpoint it had loss 7.0805 versus 6.5887 for FP8 and a
pre-clipping gradient norm of 157,696 versus 6.41.  It later exceeded 1.5
million despite gradient clipping.

The final four-arm diagnostic removed that ambiguity by crossing learned
projection format with forward P/V format while holding the backward fixed.
All four curves remain close through update 300 (78.6 million tokens).  The
MXFP4 arms visibly separate near update 400 (104.9 million tokens) and are
unambiguously split by update 500 (131.1 million tokens): loss is 7.97 for
both MXFP4 arms, compared with 5.41 for their FP8 controls.  Both MX routes
continue to fail under different learned-projection formats, whereas both FP8
routes remain non-divergent through the snapshot.  Thus forward MXFP4 P/V, or
the model state it induces, is the common separator.  This is strong
factorial evidence, but not a formal proof that one kernel instruction causes
divergence.

\begin{table}[htbp]
\centering
\caption{Initial frozen rolling-log cutoff, retained for provenance.  Jobs
began at different times, so these are status observations rather than
aligned loss or throughput comparisons.}
\label{tab:v509-four-arm-cutoff}
\small
\begin{tabular}{llrrrrl}
\toprule
Projections & Forward P/V & Update & Tokens & Loss & Grad norm & Status \\
\midrule
E4M3  & FP8   & 10,381 & 2.721B & 2.9743 & 0.2197 & working at cutoff \\
E4M3  & MXFP4 & 10,075 & 2.641B & 8.8265 & 358,400 & diverged \\
NVFP4 & FP8   & 10,884 & 2.853B & 3.1637 & 0.2051 & working at cutoff \\
NVFP4 & MXFP4 & 11,061 & 2.900B & 8.4801 & 3,915,776 & diverged \\
\bottomrule
\end{tabular}
\end{table}
\FloatBarrier

\begin{figure}[p]
\centering
\reportplot[0.34\textheight]{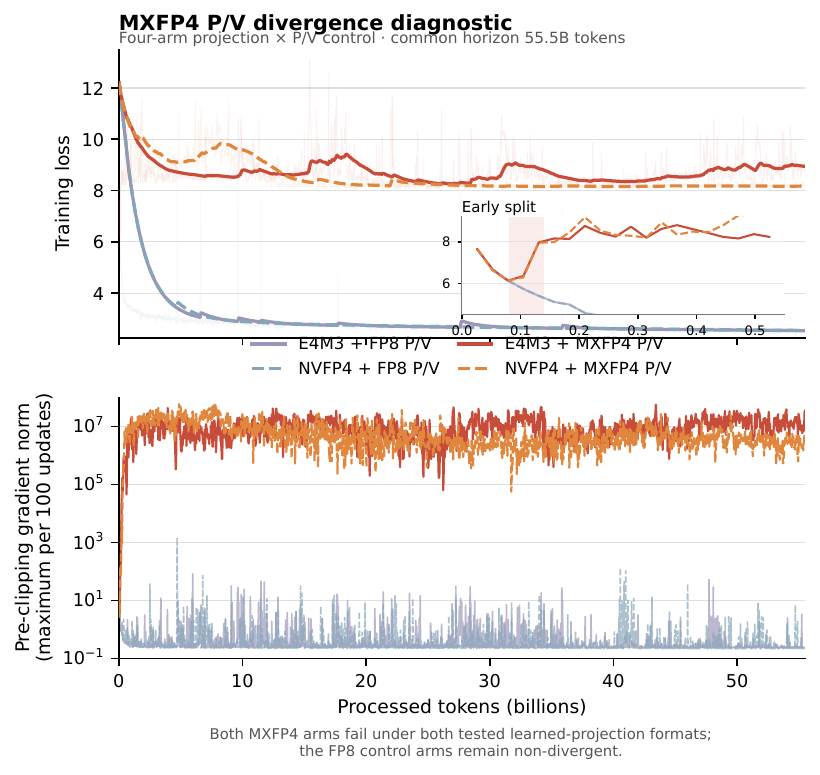}
\caption{Separated diagnostic for the two diverging MXFP4-P/V arms.  The top
panel shows loss; the lower panel shows the maximum pre-clipping gradient norm
within each 100-update display bin on a logarithmic scale.  The two FP8
control arms are muted and use the corresponding projection formats.  The
inset expands the first 0.55 billion tokens and qualitatively marks the
interval between approximately 0.08 and 0.14 billion tokens in which the
curves first separate.}
\label{fig:llama8b-mxfp4-divergence}
\end{figure}

The old rolling-log receipt first retained bad observations near update 4020
(1.054B tokens) for E4M3+MX and update 4722 (1.238B tokens) for NVFP4+MX.
The complete histories in Figure~\ref{fig:llama8b-mxfp4-divergence} show that
these were truncation markers, not onset estimates.  At the common
55.5-billion-token horizon, the one-billion-token moving-average losses are
8.94 and 8.18 for E4M3+MX and NVFP4+MX, compared with 2.53 and 2.54 for their
FP8 controls; the MX gradient norms repeatedly reach the millions.  The
receipt hashes and scientific route identities are listed in
Appendix~\ref{app:causal-training-provenance}.

\begin{figure}[p]
\centering
\reportplot[0.34\textheight]{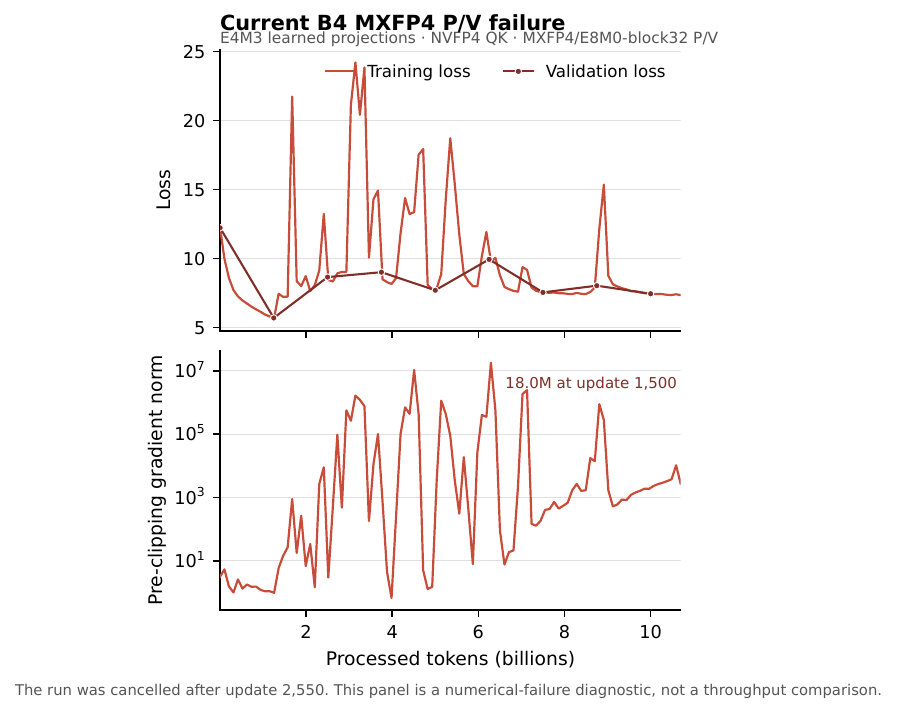}
\caption{Matched B4 MXFP4-P/V failure diagnostic.  This arm uses E4M3 learned
projections, NVFP4 Q/K inside attention, and MXFP4 P/V.  The top panel shows
training and held-out validation loss; the lower panel shows the pre-clipping
gradient norm.  The run was cancelled after update 2550 and is excluded from
the healthy-route throughput comparison.}
\label{fig:llama8b-b4-mxfp4-failure}
\end{figure}

A later B4 campaign supported the same route selection at a larger effective
batch.  The NVFP4-projection/MXFP4-PV arm diverged shortly after its initially
matched region.  A second arm changed the learned projections to E4M3 while
retaining MXFP4 P/V; it also developed repeated, finite gradient explosions
and was cancelled at update 2550.  Its last completed held-out validation was
7.4481 at update 2384.  The matched bfloat16 and
NVFP4-projection/FP8-PV trajectories remained stable.  The two failed runs
change projection precision but retain MXFP4 P/V, making the P/V
representation the common factor.  This does not prove that MXFP4 is
fundamentally unsuitable or show whether forward quantization, the saved V
payload, or its backward use is responsible.  We therefore show the current
B4 failure separately in Figure~\ref{fig:llama8b-b4-mxfp4-failure} rather than
mixing it into the healthy matched-training plot.

\FloatBarrier

\subsection{Evidence boundary}

The local brackets measure the retained backward, the complete attention
sublayer, and a full 8B update.  The distributed histories show that the FP8
route remains non-divergent and continues descending over the observed window,
while the MXFP4 route does not.  The same-recipe B4 bfloat16 control provides
same-update held-out validation and throughput over the completed
100-billion-token schedule.
Because there is one trajectory per route, these data do not support
run-to-run variability or statistical-equivalence claims.  Input stalls and
checkpoint windows remain in the reported throughput distribution.

\section{Causal-Training Result Provenance}
\label{app:causal-training-provenance}

The causal extension uses durable repository receipts rather than values
reconstructed from terminal scrollback.  The receipts below authenticate the
native attention kernels, historical transformer-step recipes, final v509
runtime, the final local timing brackets, and the four distributed
dense-cross-entropy arms.  The September 1 curves remain historical
diagnostics.  The completed September 3 receipt defines the healthy matched
B4 comparison, while the September 2 snapshot retains the separate MXFP4-P/V
failure diagnostic:

\begin{itemize}
  \item \path{results/native_tk_d64_ptx_adaptation_20260829/README.md}
  and \path{v416_llama12b_saturated_receipt_20260829.json} record the D64
  matched CuTe policies, production ABI, extension hashes, and saturated
  Llama-1.2B run.
  \item \path{results/native_tk_d128_gqa_20260829/README.md} records the
  matched v454 and v482 generated-CuTe matrices, numerical gates, resource
  envelopes, and selected-SASS identities.
  \item \path{results/tk_fa4_d128_v501_corrected_20260829/README.md}
  records the corrected statistics ABI, isolated forward result, 8B matched
  bracket, v501 profile, and straight-MX v502/v503 experiments.
  \item \path{results/tk_fa4_d128_shared_tile_mx_20260830/}
  records byte-level producer equivalence, the represented oracle, composed
  projection/backward timing, and the saturated A/control/B full-model gate.
  \item \path{results/llama8b_nvfp4_qk_backward_reconstruction_20260831/README.md}
  records the B1 native-score reconstruction and E5M2-dO diagnosis that led to
  the retained path. The final exact-batch integration identities are recorded
  by the release manifest and the timing and matched-training receipts below.
  Large tensor captures used for layer-boundary diagnosis are not
  redistributed with the paper.
  \item \path{receipts/causal_d128_report_boundaries_20260901.json}
  contains every sample and summary used for the isolated-backward and
  projection-inclusive figures.  Its SHA256 is
  \code{1dbdc27194282680\allowbreak{}7a9cb8e1b93489d3\allowbreak{}24689a77e74ad71b\allowbreak{}3deb7a85d0e47f7e};
  the authenticated raw receipt SHA256 is
  \code{0c94e8eec8bcec67\allowbreak{}68fb1e47032c5cfe\allowbreak{}e531cf73dec27c31\allowbreak{}56864ad3db8ba906}.
  The benchmark source is
  \path{tk_fa4/lowp_fa4_bwd/benchmark_v509_report_boundaries.py}.
  \item
  \path{results/tk_fa4_8b_batch_scaling_20260901/e2e_batch_scaling_summary.json}
  contains the B1/B2/B4 medians, numerical warning, compiled-extension
  identities, source commits, and hashes of the six authenticated raw timing
  receipts used by Figure~\ref{fig:llama8b-e2e-breakdown} and
  Table~\ref{tab:llama8b-e2e}.  Its SHA256 is
  \code{1849d1aed501450a\allowbreak{}4b25a807dd6855f9\allowbreak{}962f59e76e91f836\allowbreak{}848bb1a7053dab68}.
  The raw sample arrays were ephemeral and are not retained in the repository;
  the committed summary therefore supports the reported medians, not a new
  confidence-interval analysis.  All arms use standard cross entropy compiled
  by \texttt{torch.compile}; CCE is disabled.  The older
  \path{receipts/llama8b_e2e_b1_v509_20260901.json} remains as a superseded
  B1-only bracket with every sample preserved.
  \item \path{receipts/llama8b_training_curves_20260901.json} freezes the
  credential-free W\&B histories used by both training figures at
  2026-09-01 15:37:37 UTC.  Its SHA256 is
  \code{eecf78acd3cccd20\allowbreak{}f2cfae57cd8e4b9f\allowbreak{}6a79da12ecd3a453\allowbreak{}7e10f40b73591ca1}.
  Every unique token coordinate recovered from each four-arm trajectory contributes
  to its causal token-domain exponential moving average; only one row per 100
  updates is retained for display.  Duplicate coordinates created by resume
  are resolved by the final observation returned by the W\&B history scan.
  The historical reference instead retains its sampled history and committed moving average
  inside the frozen receipt; it is not treated as a matched control for the
  B4 experiment.
  \item \path{receipts/llama8b_b4_w64_launch_check_20260902.json} records the
  matched 64-GPU B4 recipe, all five aligned update-300--400 observations,
  the rejected NVFP4-projection MXFP4-P/V arm, source/runtime identities, and
  the update-239 distributed-checkpoint gate.  This launch-window receipt is
  used for route validation; aggregate throughput comes from the matched
  receipt below.  Its SHA256 is
  \code{f652ea07c34048e9\allowbreak{}180629737dc00093\allowbreak{}3e481e88856e7c64\allowbreak{}ee87f148eea21063}.
  \item The completed matched B4 histories are frozen in
  \path{receipts/llama8b_b4_completed_20260903.json}.  The receipt selects one
  checkpoint lineage per arm and contains 954 aligned training reports, 81 aligned validation
  reports, and all 874 common post-warmup throughput observations.  Terminal
  evidence records the update-23,842 checkpoint, completed remote sync, and
  successful node exit for both arms.  This receipt supports
  Figures~\ref{fig:llama8b-b4-matched-training}
  and~\ref{fig:llama8b-b4-matched-throughput}.  Its SHA256 is
  \code{36272a35bd95c313\allowbreak{}8425e7330403f94d\allowbreak{}87e40ddd2109cdcb\allowbreak{}2bcf5e2b21c1c55e}.
  \item \path{receipts/llama8b_b4_matched_snapshot_20260902T1358Z.json}
  retains the earlier matched B4 prefix together with the separate
  E4M3-projection/MXFP4-PV failure history.  The completed receipt supersedes
  its healthy-route prefix; this snapshot remains the source for
  Figure~\ref{fig:llama8b-b4-mxfp4-failure}.  Its SHA256 is
  \code{0ed4b988db3a0805\allowbreak{}d520b0d41e241d22\allowbreak{}4e0ad43e2258bf99\allowbreak{}3625d85c1af2f0da}.
\end{itemize}

The three local timing figures, two historical training figures, and three
B4 figures are regenerated from those receipts by
\path{plot_causal_training.py}.  The timing receipts also bind the
benchmark-source and compiled-runtime hashes, so the timing claims do not
depend on an unversioned extension name or terminal output.
\path{fetch_llama8b_training_curves.py} performs a read-only W\&B capture; it
requires an API key in its process environment, requires an explicit new
output path, and refuses to overwrite an existing receipt.  It never writes
the key, run configuration, or environment into the receipt.  The fetcher
validates W\&B run IDs and names; source/runtime identities are inherited from
the authenticated initial cutoff receipt rather than revalidated from W\&B.
For the completed matched B4 receipt,
\path{fetch_llama8b_b4_complete.py} reads a private source map, stitches the
selected checkpoint lineage, validates the exact token cadence, and
cross-checks four-local-rank copies in independently downloaded worker logs.
It omits service-side identifiers from its output.  The earlier MXFP4
diagnostic was frozen by \path{export_llama8b_b4_snapshot.py}.  Raw captures,
the private source map, and authentication material are not committed.

The historical distributed rows are snapshots from an earlier FP8-PV and
shared-MXFP4-PV pair.  They are used only as route-selection diagnostics, not
as checkpoint-aligned validation or aggregate throughput for the retained
recipe.
The snapshot deliberately records the matched loss gap and MX gradient
explosion so a later edit cannot silently convert an operationally running job
into a numerical-success claim.  Both jobs used learned NVFP4 QKV/O and CCE;
their values cannot be attached to the revised projection/loss contract.  No
credential-bearing configuration or environment data are included.

The initial redacted cutoff values and first-retained-instability markers are frozen in
\path{receipts/v509_four_arm_cutoff_20260831T2209Z.json}, SHA256
\code{b8e19765627e4f40\allowbreak{}d262578f9b59614f\allowbreak{}61f7eff76b35a346\allowbreak{}7b9d82b5f2784fc2}.
All four arms authenticate the same source/runtime capsules and backward
implementation and use a dense language-model head with standard cross entropy
and loss-only compilation.
Table~\ref{tab:v509-four-arm-cutoff} is the initial read-only rolling-log
cutoff from these live records.  The complete materialized histories extend
the four arms to 55.5--59.0 billion tokens and show that the old retained
markers were not onset estimates.  The two MX failures may be reported as a
factorial numerical result; the instantaneous throughput values must not be
converted into a matched speedup, the historical bfloat16 trace is not a
paired control, and the FP8 arms support stability only through their stated
common horizon.

\end{document}